\documentclass[letterpaper]{article} 
\usepackage{aaai2026}  
\usepackage{times}  
\usepackage{helvet}  
\usepackage{courier}  
\usepackage[hyphens]{url}  
\usepackage{graphicx} 
\usepackage{natbib}  
\usepackage{caption} 
\usepackage{algorithm}
\usepackage{algorithmic}

\usepackage{newfloat}

\usepackage[utf8]{inputenc} 
\usepackage[T1]{fontenc}    
\usepackage{url}            
\usepackage{booktabs}       
\usepackage{amsfonts}       
\usepackage{nicefrac}       
\usepackage{xcolor}         

\usepackage{times}
\usepackage{latexsym}
\usepackage{multirow}
\usepackage{makecell}
\usepackage{graphicx}
\usepackage{tikz}
\usepackage{pifont}
\usepackage{colortbl}
\usepackage{soul}
\usepackage{siunitx}
\usepackage{microtype}
\usepackage{enumitem}
\usepackage{inconsolata}
\usepackage{listings}
\usepackage{amsmath,amsfonts,amssymb}
\usepackage{mathtools}

\usepackage{newfloat}
\usepackage{listings}
\DeclareCaptionStyle{ruled}{labelfont=normalfont,labelsep=colon,strut=off} 
\floatstyle{ruled}
\newfloat{listing}{tb}{lst}{}
\floatname{listing}{Listing}
\title{KG-o1: Enhancing Multi-hop Question Answering in Large Language Models via Knowledge Graph Integration}
\author {
    Nan Wang\textsuperscript{\rm 1}\equalcontrib,
    Yongqi Fan\textsuperscript{\rm 1}\equalcontrib,
    yansha zhu\textsuperscript{\rm 1},
    ZongYu Wang \textsuperscript{\rm 2},
    Xuezhi Cao\textsuperscript{\rm 2}, \\
    Xinyan He\textsuperscript{\rm 2},
    Haiyun Jiang\textsuperscript{\rm 3},
    Tong Ruan\textsuperscript{\rm 1},
    Jingping Liu\textsuperscript{\rm 1}
}
\affiliations {
    \textsuperscript{\rm 1}School of Information Science and Engineering, East China University \\ of Science and Technology, Shanghai, China\\
    \textsuperscript{\rm 2}Meituan, Shanghai, China\\
    \textsuperscript{\rm 3}School of Automation and Perception, Shanghai Jiao Tong University\\
    \{y80230069, johnnyfans\}@mail.ecust.edu.cn, sanchyanprime@gmail.com, \{wangzongyu02, caoxuezhi, hexinyan03\}@meituan.com, haiyunjiangnlp@gmail.com, \{ruantong, jingpingliu\}@ecust.edu.cn
    
}

\usepackage{bibentry}

\begin{document}

\maketitle
\begin{abstract}
Large Language Models (LLMs) face challenges in knowledge-intensive reasoning tasks like classic multi-hop question answering, which involves reasoning across multiple facts. This difficulty arises because the chain-of-thoughts (CoTs) generated by LLMs in such tasks often deviate from real or a priori reasoning paths. In contrast, knowledge graphs (KGs) explicitly represent the logical connections between facts through entities and relationships. This reflects a significant gap. Meanwhile, large reasoning models (LRMs), such as o1, have demonstrated that long-step reasoning significantly enhances the performance of LLMs. Building on these insights, we propose \textbf{KG-o1}, a four-stage approach that integrates KGs to enhance the multi-hop reasoning abilities of LLMs. We first filter out initial entities and generate complex subgraphs. Secondly, we construct logical paths for subgraphs and then use knowledge graphs to build a dataset with a complex and extended brainstorming process, which trains LLMs to imitate long-term reasoning. Finally, we employ rejection sampling to generate a self-improving corpus for direct preference optimization (DPO), further refining the LLMs’ reasoning abilities. We conducted experiments on two simple and two complex datasets. The results show that KG-o1 models exhibit superior performance across all tasks compared to existing LRMs. The datasets, models, and code for KG-o1 are publicly available\footnote{\url{https://github.com/JOHNNY-fans/KG-o1#}}.
\end{abstract}

\section{Introduction}
\label{sec: Introduction}
Multi-hop question answering involves synthesizing information from multiple facts through a sequence of logical steps to derive the correct answer~\cite{trivedi2020multihop, khashabi2020unifiedqa}. For instance, answering the question \textit{``Who is the spouse of the actor who played Batman in The Dark Knight''} requires first identifying the actor who played Batman in The Dark Knight, and then determining the spouse of that actor. Although large language models have demonstrated impressive performance in question-answering tasks~\cite{kamalloo2023evaluating, wang2024no}, they still struggle to perform correct and logical reasoning in complex scenarios~\cite{kaplan2020scaling,wei2022chain,zheng2023can,ding2024break}.

Meanwhile, Knowledge Graphs (KGs) holding the potential for handling multi-hop questions, which inherently encompass logical reasoning paths between facts. For example, to answer the question above, one can follow a reasoning path between the two facts: “(\textit{The Dark Knight}, \textit{hasActor}, \textit{Christian Bale}) $\rightarrow$ (\textit{Christian Bale}, \textit{spouse}, \textit{Sibi Blazic})”, to arrive at the correct answer. Thus, in this paper, we focus on leveraging the structured nature of KGs to guide LLMs along explicit reasoning paths, thereby enhancing their performance on complex multi-hop question answering tasks~\cite{qu2020rnnlogic, choudhary2023complex}.

Although previous studies have explored leveraging KGs to enhance the multi-hop reasoning capabilities of LLMs, they have not fundamentally improved the models’ intrinsic ability to reason over multiple facts and generate reasoning processes aligned with a priori reasoning paths. Instead, KGs are typically treated as external resources. These approaches either incorporating structured data to guide chain-of-thought generation~\cite{chen2020hybridqa,zhao2024kg} or adopting retrieval-augmented generation to answer complex questions~\cite{trivedi2023interleaving,park2023graph,tang2024multihop}. Both paradigms rely on separate retrieval modules or symbolic systems, offering only limited improvements to the underlying reasoning abilities of LLMs.
Meanwhile, OpenAI's breakthrough with o1\footnote{\url{https://openai.com/o1/}} has shown that long stepwise thinking significantly improves the reasoning capabilities of LLMs~\citep{guan2024deliberative, o1p11}. LRMs employ a test-time compute scaling paradigm~\citep{snell2024testtimescaling, qi2024mutual, xi2024enhancing} to simulate slower, more deliberate Human System 2 thinking~\citep{guan2025rstar}, which improve the accuracy and flexibility of reasoning. Various methods have been proposed to in scientific, mathematical, programming, and medical reasoning tasks~\citep{zhao2024marco-o1, min2024imitate, chen2024huatuogpt, guan2025rstar, xiang2025metacot}. Despite these advances, the potential for knowledge-intensive multi-hop reasoning tasks is still underexplored.

\begin{figure}[ht]
\centering
\includegraphics[width=0.90\columnwidth]{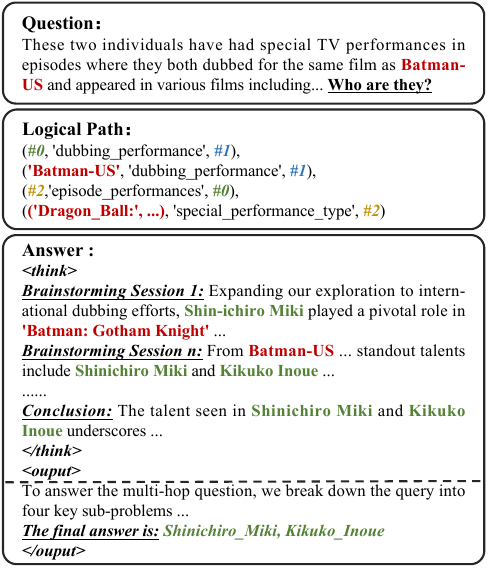} 
\caption{Example of KG-MHQA. The answer has two parts: \textit{Think}, which explicitly integrates the logical path, and \textit{Output}, which uses this reasoning to solve the multi-hop question.}
\label{fig:data sample}
\end{figure}

To this end, we propose \textbf{KG-o1}, which integrates KGs to evolve an LLM into an LRM, thereby enhancing multi-hop question answering in complex scenarios. Our approach involves extracting explicit reasoning paths from knowledge graphs and incorporating their logic into the reasoning chains of LLMs through training. As illustrated in Figure~\ref{fig:data sample}, we internalize the logical paths from KGs into the model’s intrinsic reasoning chains. The explicit process is as follows:

First, we consider relationship types to select complex subgraphs from the knowledge graph. Second, we convert these subgraphs into clear logical reasoning paths and identify entities that can be questioned. Next, the selected entities and reasoning paths are provided to ChatGPT-4o~\cite{GPT-4o} to generate multi-hop questions, resulting in complex multi-hop QA pairs. Based on these pairs, we employ data distillation to iteratively generate extended reasoning processes, yielding the KG-MHQA SFT dataset. Through supervised fine-tuning (SFT), base models are able to emulate a slow thinking paradigm, resulting in the KG-o1 SFT models. Finally, we propose the ``Self-improved Adaptive DPO'' strategy, utilizing the KG-o1 SFT models and an advanced LRM to obtain positive and negative responses through rejection sampling, thereby constructing the adaptive KG-MHQA DPO dataset. Direct preference optimization (DPO) is then applied to align the SFT models with more appropriate reasoning processes, producing the final KG-o1 models.
Our main contributions are summarized as follows:
\begin{itemize}
    \item We propose KG-o1, a framework grounded in the idea of deeply integrating KGs with LLMs. This framework provides a reusable process that systematically evolves an LLM to an LRM based on reasoning paths of KGs, thereby enhancing intrinsic multi-hop question answering for Large Language Models. 
    \item We construct the KG-MHQA SFT dataset by leveraging logical paths from the FB15k knowledge graph, and propose the Self-improve Adaptive DPO strategy to build the KG-MHQA DPO dataset. Utilizing these datasets, we develop a series of KG-o1 models.
    \item Extensive experiments on several datasets and our self-built KG-MHQA test dataset demonstrate that our KG-o1 models achieve competitive performance compared to ChatGPT-4o and o1-mini. Furthermore, ablation and domain transferability experiments verify the strong generalization ability of our approach.
\end{itemize}

\section{Related Work}
\subsection{Large Reasoning Models}
Large reasoning models enhance the performance of LLMs by applying long-term reasoning at test time, in contrast to traditional pre-trained models that prioritize scalability through increased model size or expanded training data~\cite{henighan2020scaling, yang2024qwen2, zeng2024scaling}. Recent advancements have introduced models like OpenAI-o1~\cite{jaech2024openai}, DeepSeek-R1~\cite{guo2025deepseek}, and Gemini 2.0 Flash Thinking~\cite{google2024geminiflash20}, which have undergone large-scale SFT and reinforcement learning (RL), These models demonstrate that scaling at test time can significantly improve reasoning capabilities on complex tasks such as mathematics, coding, and scientific reasoning~\cite{temsah2024openai, zhang2024o1, zhong2024evaluation}. 

Recent research has advanced the reasoning capabilities of LLMs. Open-o1~\cite{opensource_o1} and DeepSeek-R1-Distill~\cite{guo2025deepseek} leverage knowledge distillation with fine-grained filtering to provide high-quality process data for smaller LLMs. QWQ~\cite{qwq-32b-preview} and rStar Math~\cite{guan2025rstar} synthesize long reasoning via CoT, using self-evolved rewards and strategies like MCTS or recursive self-questioning to improve depth and accuracy. Recent work also refines long-form thinking quality: DRT-o1~\citep{wang2024drt} uses structured workflows for better iterative reasoning, while search-o1~\citep{li2025search} incorporates retrieval-augmented methods to bridge knowledge gaps. For training, DPO~\cite{rafailov2023direct} and PRM~\cite{lightman2023let} enhance inference via preference optimization and process supervision, while GRPO~\cite{shao2024deepseekmath} and DAPO~\cite{yu2025dapo} further improve training efficiency and long-chain inference stability.

\begin{figure*}[ht]
\centering
\includegraphics[width=\textwidth]{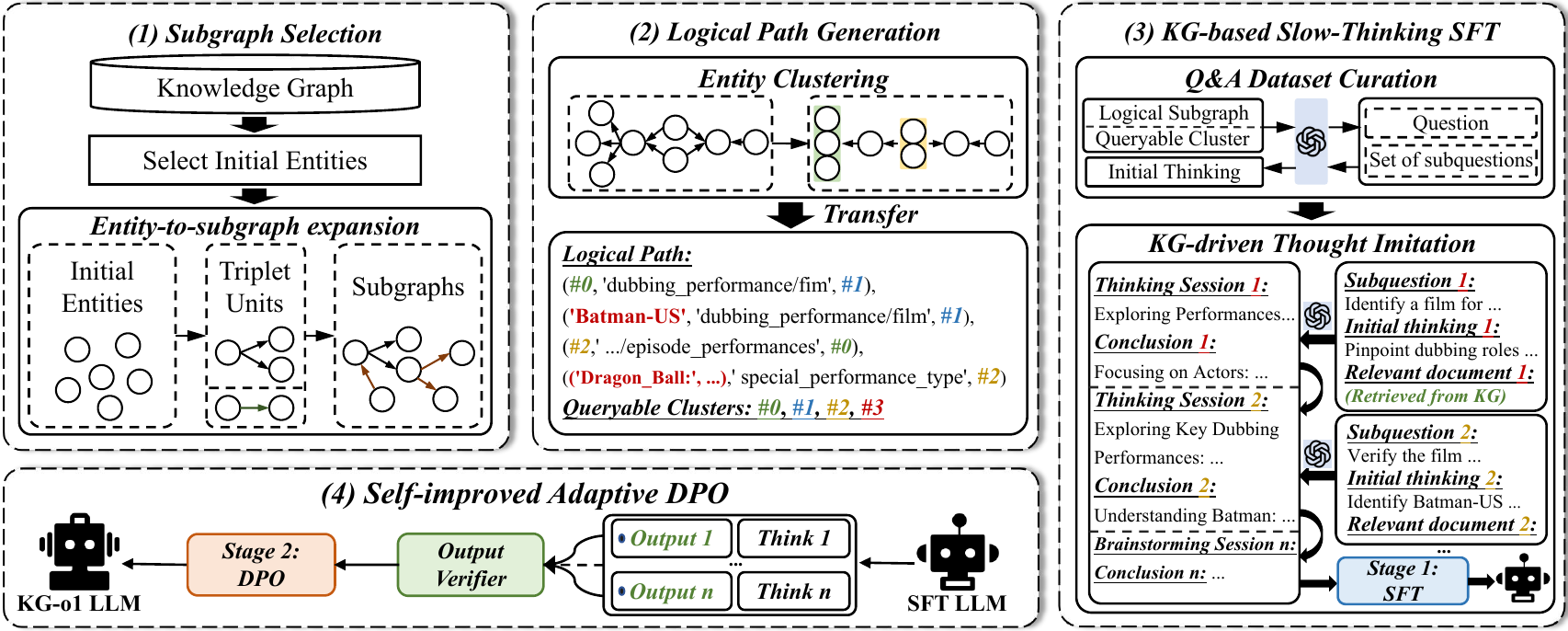} 
\caption{Overview of KG-o1 framework for enhancing multi-hop question answering in LLMs.}
\label{fig:method}
\end{figure*}

\subsection{Multi-Hop Question Answering }
Multi-hop question answering requires LLMs to reason across multiple facts. Notable datasets like HotpotQA, 2WikiMultihopQA, and MuSiQue~\cite{yang2018hotpotqa, ho2020constructing, trivedi2022musique} provide both relevant and distracting documents for robust evaluation. Recent extensions incorporate structured data, including KGs~\cite{chen2020hybridqa,mihaylov2018can,he2024mintqa} and scientific tables, while specialized domain corpora have been used in retrieval-augmented generation (RAG) systems~\cite{tang2024multihop}.

Recent advancements in LLM-based methods primarily treat KGs as external aids or use prompt engineering approaches, rather than enhancing the intrinsic reasoning abilities of LLMs. For example, CoT~\cite{saparov2023language, han2022folio} encourages step-by-step reasoning and has been adopted in methods such as PathFiD~\cite{yavuz2022modeling} and IRCoT~\cite{trivedi2023interleaving}. Similarly, question decomposition approaches~\cite{yavuz2022modeling, zhou2023leasttomost} break down complex questions into simpler sub-questions, facilitating multi-step inference. RAG-based frameworks leverage iterative cycles of retrieval and reasoning to generate multi-hop answers by combining LLMs with retrieved evidence~\cite{tang2024multihop, park2023graph}.

\section{Methodology}
\label{Methodology}
As shown in Figure~\ref{fig:method}, our method consists of four main stages. (1) Subgraph Selection: We first filter out the initial entities and then use a one-to-many approach for entity-to-subgraph expansion. (2) Logical Path Generation: Next, we transform the subgraph into logical paths and select entities that can be questioned. (3) KG-based Long-term Thinking SFT: Then, we construct complex multi-hop QA pairs, iteratively generate the SFT dataset, and fine-tune LLMs to simulate slow thinking based on the logical paths. (4) Self-improved Adaptive DPO: Finally, we generate contrastive instances and apply direct preference optimization to align models with more appropriate reasoning processes.

\subsection{SubGraph Selection}
\label{sec:preprocessing}
~\textbf{Select Initial Entities.} 
In this module, we construct the initial entity set by applying two main criteria on the knowledge graph: (1) entities must be associated with multiple distinct types to ensure cross-domain coverage, and (2) entities must be connected to a diverse set of relations, which ensures the diversity and representativeness of the final subgraphs.
In this paper, we use the widely adopted Freebase 15k knowledge graph\footnote{\url{https://huggingface.co/datasets/VLyb/FB15k}} and randomly sampled 1,000 key entities as our initial entity set for data construction.

~\textbf{Entity-to-subgraph expansion.} Given initial entities, two steps are performed to select factual subgraphs containing two and three relationship types from the knowledge graph, yielding 253,374 and 1,624,786 subgraphs, respectively.  

First, each initial entity is expanded to generate triplet units, each unit consisting of one or more triplets with a single relation type. This process is formally expressed in Formula~\ref{formula: unit}:  
\begin{equation}
\small
    \begin{aligned}
    S_{j} =  \mathbb{U}_{|R|}\{s_{r}^{e_i}\}, e_i \in N, j = 1,
    \end{aligned}
\label{formula: unit}
\end{equation}
where \( N \) is the set of initial entities, \( e_i \) is the \( i \)-th entity, \( j \) denotes the number of relation types, and \( s^{e_i} \) represents a triplet unit originates from entity \( e_i \) as its starting node. 

Second, each triplet unit is iteratively expanded by adding triplets from the knowledge graph. The process is formally expressed in Formula~\ref{formula: expansion}:  
\begin{equation}
\small
\begin{split}
    S_{j} &= S_{j-1} \cup \{ (m,r,n_i) \ \text{or} \ (n_i,r,m) | \\
    &\quad m \notin N,\ n_i \in N, \ r \in R \} , \quad j > 1
\end{split}
\label{formula: expansion}
\end{equation}
For each subgraph, all relation types \( R \) connected to its entities are identified. For each \( r \in R \), triplets are added if they satisfy: (1) one entity is in the subgraph, (2) the other is in the global knowledge graph, and (3) the connecting relation is \( r \). See Section ``Entity-to-Subgraph Expansion Examples'' in the Appendix for detailed examples.

\subsection{Logical Path Generation}
\label{sec:logic_tranfor}
In this section, we aim to transfer the subgraphs into logical paths, which are then used to select the entities that can be questioned. An explicit sample is provided in Appendix Section ``Logical Triplets Generation Examples''. Specifically, we cluster entities within each subgraph based on their connected relations:
Two entities $e_i$ and $e_j$ are grouped into the same cluster $C_k$ if, for every relation $r \in R$, they are connected to exactly the same set of entities:
$$
\forall r \in R,\ \text{Neigh}(e_i, r) = \text{Neigh}(e_j, r),
$$
where $\text{Neigh}(e, r)$ denotes the set of entities linked to $e$ by relation $r$ in either direction (i.e., as subject or object). After clustering, we transform the subgraph into logical paths by replacing each entity cluster with identifiers (e.g., ``\#1''). Finally, we apply heuristic rules to exclude clusters that are not suitable as question targets: Specifically, a cluster is excluded if its entities can be uniquely identified using only a subset of the triplets in logical paths.

\subsection{KG-based Long-term Thinking SFT}
\label{sec:KG-based Long-term Thinking Imitation}
Existing advanced LRMs have demonstrated that LLMs can remarkably enhance their reasoning ability through long-term thinking. In this section, we first prompt ChatGPT-4o (OpenAI 2024) to generate multi-hop questions, resulting in complex multi-hop QA pairs. Next, we incorporate the a priori reasoning logic and knowledge from KGs to construct detailed long-term thinking processes for each QA pair. These reasoning chains are explicitly represented and concatenated to the final response using unique tokens, resulting in the KG-MHQA SFT dataset. Through the SFT process, the models effectively simulate and exhibit long-term thinking behaviors.

~\textbf{QA Dataset Curation.} 
\label{sec: Dataset Curation}
Based on the two categories of logical paths obtained in the Logical Path Generation Section, we aim to construct multi-hop QA pairs that require complex reasoning over high-density facts. To ensure the question diversity and mitigate overfitting, we employ a uniform sampling strategy, categorizing the sampled paths based on the number of their triplets and their in-degree and out-degree patterns. This process yields comprehensive logical paths containing 2 to 6 reasoning hops, which encompass 667 distinct structural types and cover 85\% of entities. Afterward, we impose constraints regarding question targets, logical paths, and interrogative syntax to guide ChatGPT-4o~\cite{GPT-4o} in generating questions. To ensure question quality, ChatGPT-4o is further required to decompose each multi-hop question and verify that each sub-question strictly corresponds to the underlying logical path. Our prompts are exhibited in the Appendix Section ``Dataset Curation Prompts''. The corresponding entities of the question targets are used as answers, resulting in complex multi-hop QA pairs. 

~\textbf{KG-driven Thought Imitation.} 
At this stage, we aim to construct KG-based long-term thinking SFT data for fine-tuning LLMs to imitate the long-term thinking paradigm. Specifically, we integrate KGs and employ an iterative approach to construct the long-term thinking process. Given the complex multi-hop QA pairs, we plan the iterative optimization sequence according to prior reasoning paths to ensure that the process adheres to the reasoning logic.
To ensure that the reasoning process is guided by the provided knowledge graph, we retrieve relevant triples at each iteration, present them in textual form, and provide them to the large language model as references. This approach encourages the model to integrate internal knowledge with external information, facilitating a more thorough and comprehensive reasoning process.
The iterative procedure is formally defined in Formula~\ref{formula:inter_ref}:
\begin{equation}
\small
    \begin{aligned}
    P_n = f(Q_n, T_n, K_n, P_1, P_2, \dots, P_{n-1}),
    \end{aligned}
\label{formula:inter_ref}
\end{equation}
where \( P_n \) is the refined thinking process at the \( n \)-th iteration, \( Q_n \) is the subquestion, \( T_n \) is corresponding initial thinking segment, and \( K_n \) is the knowledge for the current iteration. The previous iterations \( P_1, P_2, \dots, P_{n-1} \) are used to refine the thinking in the current iteration.

After generating the long-term thinking process, we introduce special tokens to distinguish the thought process from the output. In the \textit{think} part, the model engages in brainstorming and extensive reasoning, while in the \textit{output} part, it summarizes the reasoning steps, derives a solution from the subquestions, and generates the final answer. Using these QA pairs, we construct the KG-MHQA SFT dataset and fine-tune LLMs, enabling them to effectively emulate long-term reasoning. The sample data and statistics of KG-MHQA SFT dataset are shown in Figure~\ref{fig:data sample} and Table~\ref{tab:dataset_split_hops}.

\subsection{Self-improved Adaptive DPO}
\label{sec: Self-improved Adaptive DPO}
Direct Preference Optimization (DPO) has been demonstrated as an effective approach for reducing harmful outputs and improving generation quality in large language models. To improve the long-term thinking process quality of our KG-o1 SFT models, we implement DPO optimization after the supervised fine-tuning stage. Due to variations in training data and model architectures, the knowledge and reasoning capabilities of different models vary, making SFT data distilled from a single source LLM insufficient to achieve consistent improvements across models.

To address this limitation, we propose a ``Self-improved Adaptive DPO'' strategy that dynamically constructs contrastive positive and negative pairs by combining SFT data with KG-o1 SFT model responses. As shown in stage 2 of Figure~\ref{fig:method}, for each sample in the SFT dataset, we implement a dual path contrastive data construction protocol using the correctness of the answers as the gold standard: 
When the response generated by KG-o1 SFT model matches the correct answer, it is treated as a positive instance, with the corresponding SFT sample response as the negative. If the generated response is incorrect, the original SFT sample response remains positive, while the chain-of-thought response generated by ChatGPT-4o is used as the negative. This adaptive strategy ensures stable alignment of reasoning processes and response quality. The resulting KG-MHQA DPO dataset is then used for DPO optimization on the KG-o1 SFT model, aligning suitable long-term thinking processes for multi-hop question answering.

\begin{table}[t]
    \centering
    \fontsize{9}{10}\selectfont
    \setlength{\tabcolsep}{1.9mm}
    \begin{tabular}{lccccccc}
        \toprule
        \textbf{Split} & \textbf{Total} & \textbf{2-Hop} & \textbf{3-Hop} & \textbf{4-Hop} & \textbf{5-Hop} & \textbf{6-Hop} \\
        \midrule
        Train & 5030 & 576 & 771 & 938 & 1,416 & 1,329 \\
        Dev   & 456  & 103  & 110 & 115  & 113  & 15 \\
        Test  & 307  & 51 & 54  & 73  & 104  & 25 \\
        \bottomrule
    \end{tabular}
    \caption{Statistics of KG-MHQA SFT dataset.}
    \label{tab:dataset_split_hops}
\end{table}

\begin{table*}[t!]
    \centering
    \fontsize{9}{10}\selectfont
    \setlength{\tabcolsep}{0.85mm}
    \begin{tabular}{cccccccccccccccccc}
        \toprule
        \multicolumn{1}{c}{\multirow{2}*{\textbf{Model}}} & \multicolumn{1}{c}{\multirow{2}*{\textbf{Method}}} & \multicolumn{4}{c}{\textbf{KG-MHQA $\dagger$}} & \multicolumn{4}{c}{\textbf{2WikiMultiHopQA}} & \multicolumn{4}{c}{\textbf{HotpotQA}} & \multicolumn{4}{c}{\textbf{MINTQA}} \\
        \cmidrule(r){3-6} \cmidrule(r){7-10} \cmidrule(r){11-14} \cmidrule(r){15-18}
         & & \textbf{EM} & \textbf{F1} & \textbf{Prec.} & \textbf{Rec.} & \textbf{EM} & \textbf{F1} & \textbf{Prec.} & \textbf{Rec.} & \textbf{EM} & \textbf{F1} & \textbf{Prec.} & \textbf{Rec.} & \textbf{EM} & \textbf{F1} & \textbf{Prec.} & \textbf{Rec.} \\
        \midrule
        \multicolumn{18}{c}{\textit{\textbf{Large Reasoning Models}}} \\
        \midrule
        o1-mini & CoT & 41.69 & 53.25 & 51.77 & 60.33 & 42.60 & 61.46 & 56.35 & 78.90 & 49.80 & 67.33 & 67.47 & 76.67 & 75.67 & 85.96 & 85.20 & 90.21 \\
        Open-o1 & CoT & 44.33 & 60.33 & 56.36 & \textbf{79.46} & 25.08 & 36.64 & 35.04 & 44.80 & 35.20 & 55.13 & 49.77 & 72.06 & 37.60 & 55.05 & 53.75 & 69.78  \\
        QwQ-32B & CoT & 38.44 & 49.04 & 47.45 & 64.20 & \textbf{68.33} & \textbf{79.06} & \textbf{77.92} & 84.62 & 56.20 & 70.96 & 67.70 & \textbf{82.75} & 48.40 & 64.94 & 67.04 & 68.94 \\
        Gemini-2.0-FT & CoT & \textbf{48.21} & \textbf{60.40} & \textbf{57.76} & 70.23 & 66.20 & 76.23 & 74.13 & 82.84 & \textbf{57.00} & \textbf{72.88} & 74.95 & 79.60 & 80.67 & 88.53 & 87.49 & \textbf{92.28} \\
        DeepSeek-R1 & CoT & 44.95 & 56.47 & 54.79 & 66.99 & 67.00 & 76.76 & 75.30 & \textbf{86.33} & \textbf{57.00} & 72.24 & \textbf{75.21} & 78.34 & \textbf{85.67} & \textbf{88.95} & \textbf{88.80} & 91.21 \\
        \midrule
        \multicolumn{18}{c}{\textit{\textbf{General Purpose Large Language Models}}} \\
        \midrule
        ChatGPT-4o-mini & CoT & 33.22 & 42.94 & 42.04 & 48.55 & 62.40 & 72.93 & 71.50 & 78.02 & 53.20 & 69.50 & 71.73 & 71.47 & 78.00 & 83.24 & 82.96 & 84.96 \\
        ChatGPT-4o & CoT & \textbf{42.02} & \textbf{55.53} & \textbf{53.10} & \textbf{65.88} & \textbf{70.00} & \textbf{81.02} & \textbf{78.73} & \textbf{86.95} & \textbf{56.40} & \textbf{72.56} & \textbf{74.66} & \textbf{77.81} & \textbf{78.33} & \textbf{87.42} & \textbf{86.21} & \textbf{90.94} \\
        \midrule
        \multirow{2}{*}{Llama3.1-8B} & CoT & 23.45 & 36.50 & 34.78 & 56.97 & 36.59 & 46.84 & 45.40 & 56.59 & 42.00 & 55.54 & 56.98 & 65.94 & 54.00 & 71.26 & 68.84 & 82.68 \\
         & Ours & 41.04 & 57.81 & 53.63 & 69.44 & 55.00 & 68.59 & 66.84 & 73.77 & 43.40 & 60.16 & 62.20 & 62.24 & 67.00 & 75.97 & 74.92 & 79.02 \\ 
        \multirow{2}{*}{Qwen2.5-7B} & CoT & 14.98 & 24.34 & 22.86 & 48.57 & 29.00 & 44.50 & 40.17 & 70.07 & 13.35 & 25.90 & 23.39 & 75.51 & 21.67 & 33.33 & 29.58 & 85.50 \\
         & Ours & 30.29 & 43.59 & 41.22 & 61.12 & 32.40 & 41.00 & 39.72 & 59.00 & 25.40 & 36.35 & 37.67 & 54.76 & 50.00 & 58.08 & 57.05 & 75.26 \\
        \multirow{2}{*}{DeepSeek-14B} & CoT & 37.78 & 48.09 & 46.64 & 54.96 & 54.20 & 68.91 & 66.26 & 77.46 & 48.02 & 64.62 & 66.70 & 67.81 & 69.23 & 79.06 & 78.78 & 81.99 \\
        & Ours & 34.31 & 45.81 & 43.78 & 57.73 & 45.89 & 62.86 & 59.23 & 74.98 & 44.17 & 62.25 & 62.10 & 72.49 & 64.66 & 75.67 & 74.60 & 84.99 \\
        \multirow{2}{*}{Qwen2.5-14B} & CoT & 29.97 & 38.97 & 37.64 & 54.78 & 57.20 & 68.15 & 65.86 & 79.44 & 42.20 & 55.31 & 55.14 & \textbf{78.66} & 61.00 & 70.50 & 68.77 & \textbf{89.18} \\
         & Ours & \textbf{45.60} & \textbf{69.36} & \textbf{63.75} & \textbf{84.46} & \textbf{62.40} & \textbf{74.45} & \textbf{72.56} & \textbf{79.45} & \textbf{51.00} & \textbf{67.11} & \textbf{69.66} & 68.18 & \textbf{72.33} & \textbf{79.37} & \textbf{78.73} & 81.49  \\
        \bottomrule
    \end{tabular}
    \caption{Exact-Match (EM), F1, Precision, and Recall results of Large Reasoning Models and General Purpose Large Language Models on multi-hop reasoning datasets. $\dagger$ represents our self-build dataset, Gemini-2.0-FT denotes Gemini-2.0-Flash-Thinking-exp, DeepSeek-14B denotes DeepSeek-R1-Distill-Qwen-14B.}
    \label{tab:main}
\end{table*}

\section{Experiments}
In this section, we evaluate the performance of prominent LRMs, advanced GPLLMs, and our KG-o1 models on four datasets. Afterwards we provide a detailed analysis.

\subsection{Datasets}
\label{Datasets}
We conduct our main experiments on three open-source datasets: HotpotQA, 2WikiMultiHopQA, and MINTQA as well as our self-constructed KG-MHQA test dataset. Dataset statistics are shown in Table~\ref{tab: datasets}. For HotpotQA and 2WikiMultiHopQA, we follow the standard experimental setup of previous work~\citep{trivedi2023interleaving}, using 500 sampled instances for each dataset. For MINTQA~\cite{he2024mintqa}, which focuses on new and tail knowledge, we remove questions whose entities overlap with the KG-MHQA SFT training data to prevent knowledge leakage, and then uniformly sample 300 questions requiring 2–4 reasoning hops.

For our self-built KG-MHQA test set, we conduct entity-based partitioning and rigorous manual annotation to ensure data quality, resulting in a test set of 307 high-quality samples. The annotation team, comprising one PhD student and one Master's student with NLP expertise, independently verified both correctness and reasoning coherence of QA pairs, resolving discrepancies through group discussion. The inter-annotator agreement report is in Section ``Inter-annotator agreement report'' in the Appendix.

\begin{table}[ht]
    \centering
    \fontsize{9}{10}\selectfont
    \setlength{\tabcolsep}{1.3mm}
    \begin{tabular}{lccccccc}
        \toprule
        \multirow{1}{*}{\textbf{Datasets}} & \multicolumn{1}{c}{\textbf{\# Hop}} & \multicolumn{1}{c}{\textbf{\# Test Samples}} & \multirow{1}{*}{\textbf{\# Avg. Doc. Len.}}\\ 
        \midrule
        HotpotQA   & 2  & 500 & 1791.61 \\
        2WikiMultiHopQA   & 2,4 & 500  & 984.52 \\
        MINTQA  & 2\text{ to }4 & 300 & 1006.33 \\
        KG-MHQA & 2\text{ to }6 & 307 & 1094.45 \\
        \bottomrule
    \end{tabular}
    \caption{Statistics and description of test datasets.}
    \label{tab: datasets}
\end{table}

\begin{figure*}[t]
\centering
\includegraphics[width=\textwidth]{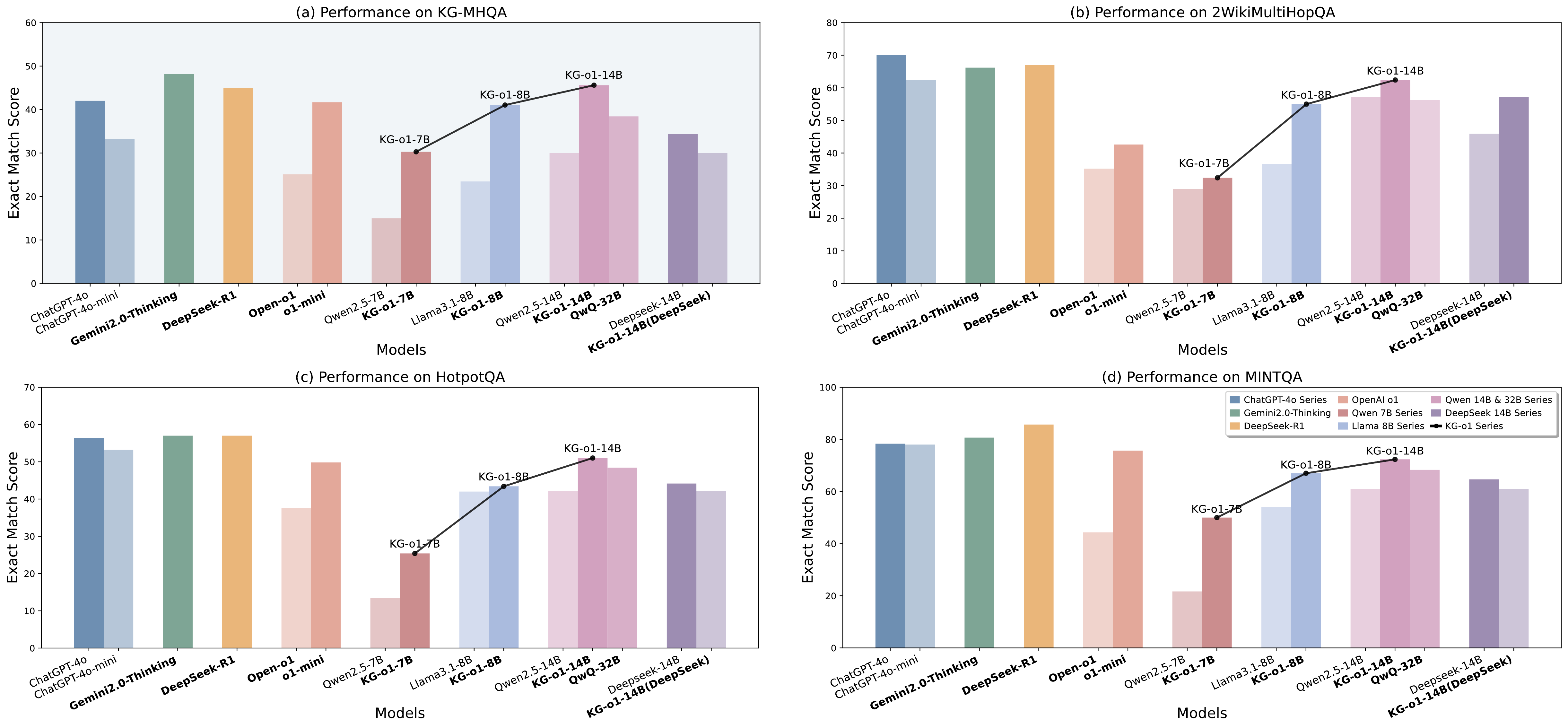}{
\caption{Comparison of model size and performance for GPLLMs, LRMs, and KG-o1 models on four multi-hop reasoning datasets using Exact Match as a metric. Long-term thinking models are indicated in bold.}
\label{fig: Main_result_2}
}
\end{figure*}

\subsection{Baseline Models}
We adopt the Qwen2.5-14B-Instruct, Qwen2.5-7B-Instruct, Llama3.1-8B-Instruct, and DeepSeek-R1-Distill-Qwen-14B  as our backbones. The primary experiments focus on Qwen2.5-14B-Instruct. For comparison, we select several prominent LRMs, including o1-mini, DeepSeek-R1, QwQ-32B-Preview, Gemini-2.0-Flash-Thinking, and an influential model, Open-o1, in the open source community. Additionally, we include two advanced general-purpose large language models (GPLLMs): ChatGPT-4o and ChatGPT4o-mini.  

\textbf{Implementations and Metrics}: 
In our experiments, we train our KG-o1 models on the KG-MHQA training set through SFT and DPO. Our training details are shown in Appendix Section ``Training Details''. For evaluation, we report exact match (EM), F1 score (F1), precision (Prec.), and recall (Rec.) results of baseline models in four datasets. The settings of evaluation are shown in Appendix Section ``Evaluation Metrics and Hyper-parameter Setting''.

\subsection{Main Results}
\label{sec:analysis}
We compare the KG-o1 models with several baselines, including both SFT and preference learning approaches, as well as advanced LRMs, powerful GPLLMs, and basic LLMs. The results are presented in Table~\ref{tab:main}. Our experiments demonstrate that advanced LRMs perform well in multi-hop question answering, with Deepseek-R1 and Gemini-2.0-Flash-Thinking-exp leading the pack, outperforming the other LRMs by a wide margin. Meanwhile, preference learning methods further enhance the reasoning capabilities of open-source models, both our KG-o1 and GRPO training method substantially improve the reasoning performance. Notably, GPLLMs achieve strong results without employing the long-term thinking paradigm, even surpassing Deepseek-R1 on 2WikiMultiHopQA, indicating that advanced GPLLMs can excel with traditional CoT. When comparing KG-o1 models with their base models, our method significantly improves performance across the three open-source and KG-MHQA test datasets, highlighting both its effectiveness and generalization ability.

Next, we analyze LLMs' multi-hop reasoning ability through two dimensions: knowledge ability and reasoning ability. Since this ability cannot be directly decomposed, we link knowledge ability to parameter scale and reasoning ability to reasoning paradigms. Figure~\ref{fig: Main_result_2} visualizes performance variations across model series.

\textbf{Analyzed from knowledge capability}. As model parameters increase, the performance of LLMs within the same series and reasoning paradigm improves significantly. For instance, the performance boost from the 7B to 14B parameter scale in the Qwen series is substantial. However, once the number of parameters reaches a certain threshold, further increases yield diminishing returns in knowledge capacity. For example, on the MINTQA dataset, the performance gain between the 4o-mini and 4o models is only 0.33, with a similar trend observed across the other three datasets. This suggests that as model size grows, the marginal improvement in knowledge capacity and reasoning ability declines.

\textbf{Analyzed from reasoning ability}. Comparing LLMs with different inference patterns at the same parameter scale reveals that the long-term thinking model exhibits superior multi-hop reasoning ability. Specifically, experiments comparing our trained KG-o1 model with base models (e.g., KG-o1-7B vs. Qwen2.5-7B, KG-o1-14B vs. Qwen2.5-14B) show significant performance improvements. To further explore the relationship between reasoning ability and patterns, we conducted ablation studies on four datasets. As shown in (a) in Figure~\ref{fig: Main_result_2}, the KG-o1 model outperforms the base model after training, indicating that the long-term thinking paradigm enhances multi-hop reasoning. Notably, even when knowledge ability improvements are similar (as seen from the comparable slopes in (a) in Figure~\ref{fig: Main_result_2}), the reasoning ability improvement is substantially greater than in the base model. This finding is further validated across three generalized experimental datasets, where both KG-o1-7B and KG-o1-14B show consistent performance gains, demonstrating the effectiveness of our approach in enhancing multi-hop reasoning.

\begin{table*}[t]
    \centering
    \setlength{\tabcolsep}{1.3mm} 
    \fontsize{9}{10}\selectfont
    \begin{tabular}{lcccccccccccccccc}
        \toprule
        \multicolumn{1}{c}{\multirow{2}*{\textbf{Method}}} & \multicolumn{4}{c}{\textbf{KG-MHQA $\dagger$}} & \multicolumn{4}{c}{\textbf{2WikiMultiHopQA}} & \multicolumn{4}{c}{\textbf{HotpotQA}}  & \multicolumn{4}{c}{\textbf{MINTQA}} \\
        \cmidrule(r){2-5} \cmidrule(r){6-9} \cmidrule(r){10-13} \cmidrule(r){14-17}
        & \textbf{EM} & \textbf{F1} & \textbf{Prec.} & \textbf{Rec.} & \textbf{EM} & \textbf{F1} & \textbf{Prec.} & \textbf{Rec.} & \textbf{EM} & \textbf{F1} & \textbf{Prec.} & \textbf{Rec.} & \textbf{EM} & \textbf{F1} & \textbf{Prec.} & \textbf{Rec.} \\
        \midrule
        CoT & 29.97 & 38.97 & 37.64 & 54.78 & 57.20 & 68.15 & 65.86 & 79.44 & 42.20 & 55.31 & 55.14 & \textbf{78.66} & 61.00 & 70.50 & 68.77 & \textbf{89.18}   \\
        SFT (QA) & 43.65 & 65.22 & 60.05 & 78.42 & 36.80 & 54.15 & 50.83 & 62.47 & 42.40 & 60.19 & 61.34 & 64.18 & 60.00 & 76.86 & 72.84 & 88.58 \\
        SFT & 41.37 & 58.89 & 54.92 & 69.27 & 59.20 & 72.25 & 70.47 & 77.45 & 48.60 & 66.67 & 68.85 & 68.32 & 68.33 & 76.89 & 75.84 & 80.77  \\
        DPO & 30.07 & 39.98 & 38.38 & 52.45 & 46.89 & 62.91 & 59.37 & 74.65 & 42.37 & 59.81 & 60.31 & 70.00 & 66.67 & 78.14 & 76.62 & 86.43  \\
        GRPO & 34.96 & 45.71 & 43.91 & 54.93 & 51.50 & 66.47 & 63.25 & 76.58 & 47.59 & 64.84 & 65.98 & 69.36 & 67.33 & 78.51 & 77.35 & 86.85 \\
        SADPO & 34.31 & 45.81 & 43.79 & 57.73 & 45.89 & 62.87 & 59.23 & 74.98 & 44.18 & 62.25 & 62.11 & 72.50 & 64.67 & 75.67 & 74.60 & 85.00  \\
        SFT + DPO & 41.37 & 59.12 & 55.14 & 69.58 & 61.00 & 73.09 & 71.33 & 78.07 & 48.80 & 65.72 & 67.75 & 67.53 & 69.67 & 76.59 & 75.69 & 79.26 \\
        SFT + GRPO & 43.13 & 59.62 & 55.68 & 69.68 & 61.01 & 72.59 & 70.90 & 7729 & 48.39 & 65.75 & 67.72 & 68.13 & 71.61 & 78.13 & 77.60 & 80.35\\
        SFT + SADPO & \textbf{45.60} & \textbf{69.36} & \textbf{63.75} & \textbf{84.46} & \textbf{62.40} & \textbf{74.45} & \textbf{72.56} & \textbf{79.45} & \textbf{51.00} & \textbf{67.11} & \textbf{69.66} & 68.18 & \textbf{72.33} & \textbf{79.37} & \textbf{78.73} & 81.49 \\
        \bottomrule
    \end{tabular}
    \caption{Exact-Match (EM), F1, Precision, and Recall results of ablation study of Qwen2.5-14B-Instruct on multi-hop question answering datasets. $\dagger$ represents our self-build dataset, SADPO denotes Self-improved Adaptive DPO.}
    \label{tab:ablation}
\end{table*}

\subsection{Ablation Study}
In our approach, we systematically conduct ablation experiments for different training phases and reinforcement learning methods, the results are presented in Table~\ref{tab:ablation}. In addition, we specifically designed the ablation experiments for medical domain transferability.

\textbf{SFT LLM (QA) vs. CoT:} SFT LLM (QA) is the model training only on the QA dataset. The QA model shows limited improvement or even a decrease in performance compared to the CoT approach, e.g., on the 2Wikimultihopqa dataset, the performance decreases to 36.80\%. 

\textbf{SFT LLM (QA) vs. SFT LLM:} SFT LLM is trained solely on the KG-MHQA SFT dataset. Comparing the SFT LLM with the QA LLM reveals that while both models incorporate knowledge from the training set, the long-thinking model can effectively analyze the question, plan the solution, integrate knowledge into the reasoning chain, and iteratively validate it to arrive at the final answer.

\textbf{RL vs. SFT:} 
Experimental results indicate that reinforcement learning (RL) surpasses supervised fine-tuning (SFT) in boosting LLMs' multi-hop reasoning. Notably, Self-improved Adapative DPO outperforms GRPO in guiding LLMs to produce coherent long-term reasoning and accurate answers, which stems from its adaptive self-improvement and preference-adjustment mechanisms.

\textbf{Domain Transferability:} We choose the MedQA-MedGENIE test set~\citep{frisoni2024generate} for domain transferability experiment. We use GPT-4o to reformat the multiple-choice questions to generative questions and use the provided contexts as the input knowledge. Experimental result in Table~\ref{tab:medqa_results} shows that GPLLMs, such as ChatGPT-4o, demonstrate superior performance over LRM on the medical reasoning dataset. Meanwhile, our proposed method (trained solely on a general knowledge graph) significantly improves the medical reasoning ability of the base model, achieving competitive performance comparable to DeepSeek-R1. This validates the effectiveness of our methods in cross-domain transferability.

\begin{table}[ht]
    \centering
    \fontsize{9}{10}\selectfont
    \setlength{\tabcolsep}{1.7mm} 
    \begin{tabular}{lccccc}
        \toprule
        \textbf{Model} & \textbf{Method} & \textbf{EM} & \textbf{F1} & \textbf{Prec.} & \textbf{Rec.} \\
        \midrule
        ChatGPT-4o-mini & CoT & 20.68 & 38.73 & 36.81 & 47.54 \\
        ChatGPT-4o & CoT & \textbf{25.24} & \textbf{44.33} & \textbf{42.78} & \textbf{51.47} \\
        Qwen2.5-14B & CoT & 13.87 & 28.62 & 26.16 & 46.33 \\
        \midrule
        DeepSeek-R1 & Reasoning & 17.49 & 27.85 & 27.64 & 32.52 \\
        DeepSeek-14B & Reasoning & 16.65 & \textbf{33.64} & 30.79 & \textbf{46.62} \\
        KG-o1-14B & Reasoning & \textbf{17.99} & 32.97 & \textbf{33.66} & 35.80 \\
        \bottomrule
    \end{tabular}
    \caption{Performance on MedQA-MedGENIE}
    \label{tab:medqa_results}
\end{table}


\subsection{Error and Cost Analysis}
\begin{figure}[ht]
    \centering
    \includegraphics[width=1.0\columnwidth]{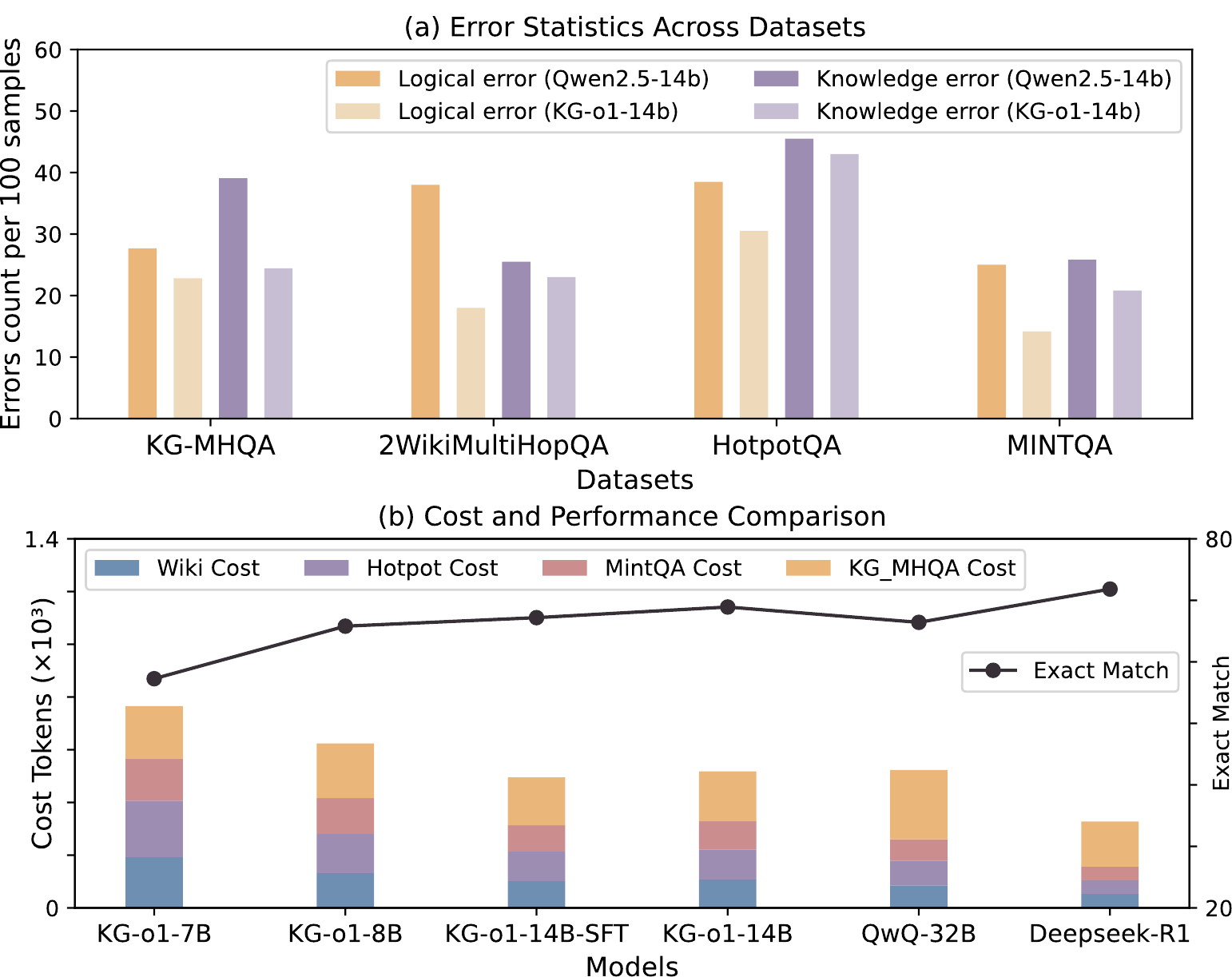}
    \caption{(a) Error statistics for Qwen2.5-14B-Instruct and KG-o1-14B models on four datasets. (b) Cost and overall performance of six LRMs
across four datasets}
    \label{fig:Error_statistics}
\end{figure}

Figure~\ref{fig:Error_statistics} summarizes both error analysis and resource-performance trade-offs for KG-o1 and baseline models across four datasets. In (a), we manually classify 100 randomly sampled errors into knowledge errors, where the model uses irrelevant or incorrect knowledge, and logic errors, which result from faulty multi-hop reasoning. Results show that our approach substantially reduces knowledge errors on KG-MHQA and logic errors on the other datasets, especially 2WikiMultiHopQA, demonstrating the effectiveness of our approach (see Appendix Section “Cases of Two Error Types” for details). In (b), as the size of the model parameter increases, the KG-o1 models generate shorter outputs, consume fewer tokens, and achieve better performance, a trend also seen in QWQ-32B-preview and Deepseek-R1. Furthermore, KG-o1 models display more consistent token usage compared to Deepseek-R1, likely because our training set’s long reasoning sequences encourage stable response lengths. 

\section{Conclusion and Limitations}
In conclusion, we propose KG-o1, which integrates KGs to evolve an LLM to an LRM for complex knowledge-intensive multi-hop question answering. We first select complex fact subgraphs from KGs and generate logical paths. Further, we construct complex multi-hop QA pairs and iteratively generate the KG-MHQA SFT dataset for KG-driven long-term thinking SFT. Additionally, we construct the KG-MHQA test dataset for model evaluation. Finally, we propose the ``Self-improved Adaptive DPO'' strategy to obtain the adaptive KG-MHQA DPO dataset for the alignment of appropriate reasoning. We then conducted experiments on four multi-hop reasoning datasets and a domain transferability dataset of advanced LRMs, GPLLMs, and our KG-o1 models, further validating the effectiveness of our proposed method.

Although we have implemented strategies to enhance the quality of the thinking process, our current approach primarily treats the thinking process as a whole for contrastive learning. More granular alignment of the model's thinking content remains space for future exploration. Furthermore, when constructing multi-hop questions, we focused on free-text answers, overlooking other answer formats such as comparisons or yes/no judgments. These limitations highlight potential directions for improvement, and we will continue to explore ways to refine and advance our work.

\section{Ethical Statement}
\label{Ethical Considerations}
This paper proposes KG-o1, which integrates KGs to enhance the multi-hop reasoning abilities of LLMs. We ensured ethical compliance by using publicly available or copyright-compliant datasets and avoiding sensitive information. Our model prioritizes factual and logical consistency to reduce misinformation risks, and our methods are transparently documented for reproducibility. We also optimized computational resources to reduce environmental impact, aligning our work with ethical AI research standards. We recognize that applying KG-o1 to reasoning tasks in diverse real-world scenarios may have broader impacts. To prevent potential misuse or misinterpretation, we recommend users validate model outputs with relevant experts before adopting them for important decisions, ensuring responsible and ethical application.

\bibliography{aaai2026}

\clearpage

\appendix

\section{Access of KG-o1 approach}
\label{app:github}
We provide an anonymous link to a GitHub repository containing \textbf{SFT and DPO datasets}, \textbf{training hyperparameters}, \textbf{evaluation results}, and \textbf{KG-o1 models}. Meanwhile, we have added an open-source license to the project.

\textbf{URL:}~\url{https://anonymous.4open.science/r/KG-o1-F493}

\section{Entity-to-Subgraph Expansion Examples.}
\label{sec: Entity-to-subgraph expansion}
\subsection*{Step 1: Initial Triplet Units Construction}

\textbf{Formula Definition}
\begin{equation}
\begin{aligned}
S_{j} = \mathbb{U}_{|R|}\{s_{r}^{e_i}\}, e_i \in N, j = 1
\end{aligned}
\end{equation}

\textbf{Initial Entity:} Batman:\_Gotham\_Knight

\textbf{Selected Relation Type:}

\texttt{/film/dubbing\_performance/film}

\textbf{Generated Unit:}
\begin{verbatim}
[
  ["Shinichiro_Mik", "/film/dubbing_performance/
  film", "Batman:_Gotham_Knight"],
  ["Kikuko_Inoue", "/film/dubbing_performance/
  film", "Batman:_Gotham_Knight"],
  ["Batman-US", "/film/dubbing_performance/film",
  "Batman:_Gotham_Knight"]
]
\end{verbatim}

In this example, $N$ is the set of initial entities, while \texttt{Batman:\_Gotham\_Knight} represents the explicit entity $e_i$ in $N$. Select a relation type $r$ connected to $e_i$, i.e., \texttt{/film/dubbing\_performance/film}, to expand the entities (Shinichiro\_Mik, Kikuko\_Inoue, Batman-US) to get a unit. $\mathbb{U}$ is the set of all units.

\subsection*{Step 2: Triplets Expansion}

\textbf{Formula Definition}
\begin{equation}
\small
\begin{split}
    S_{j} &= S_{j-1} \cup \{ (m,r,n_i) \ \text{or} \ (n_i,r,m) | \\
    &\quad m \notin N,\ n_i \in N, \ r \in R \} , \quad j > 1
\end{split}
\label{formula: expansion}
\end{equation}

\textbf{Selected Relation Type:} 

\texttt{/tv/special\_tv\_performance\_type/episode\_performances}

\textbf{Expanded Subgraph:}
\begin{verbatim}
[
  ["Shinichiro_Mik", "/film/dubbing_performance/
  film", "Batman:_Gotham_Knight"],
  ["Kikuko_Inoue", "/film/dubbing_performance/
  film", "Batman:_Gotham_Knight"],
  ["Batman-US", "/film/dubbing_performance/film",
  "Batman:_Gotham_Knight"],
  ["Seiyū-GB", "/tv/special_tv_performance_type/
  episode_performances", "Shinichiro_Mik"],
  ["Seiyū-GB", "/tv/special_tv_performance_type/
  episode_performances", "Kikuko_Inoue"]
]
\end{verbatim}

$R$ represents the concatenated set of all relational types, each relational type $r$ is connected with one or more nodes in $N$. In this example, $r$ is \texttt{/tv/special\_tv\_performance\_type/episode\_performances}. 

Next, we need to find an entity connected to $r$ and contained in $N$ as $n_i$. Locate the triplets in KG and add them to the unit. The triplets should satisfy that the relationship type is $r$ and one of the head or tail entities is $n_i$. For example, two triplets were added in this round of extensions: 

\begin{verbatim}
[
  ["Seiyū-GB", "/tv/special_tv_performance_type/
  episode_performances", "Shinichiro_Mik"],
  ["Seiyū-GB", "/tv/special_tv_performance_type/
  episode_performances", "Kikuko_Inoue"]
]
\end{verbatim}

\section{Logical Triplets Generation Examples.}
\label{sec: Logical Triplets Generation}
\subsection*{Step 1: Cluster Triplet Formation}

\textbf{Original Triplets:}
\begin{verbatim}
[
  ["Shinichiro_Mik", "/film/dubbing_performance/
  film", "Batman:_Gotham_Knight"],
  ["Kikuko_Inoue", "/film/dubbing_performance/
  film", "Batman:_Gotham_Knight"],
  ["Batman-US", "/film/dubbing_performance/film",
  "Batman:_Gotham_Knight"],
  ["Seiyū-GB", "/tv/special_tv_performance_type/
  episode_performances", "Shinichiro_Mik"],
  ["Seiyū-GB", "/tv/special_tv_performance_type/
  episode_performances", "Kikuko_Inoue"],
  ["Sailor_Moon_S_the_Movie", "/film/performance/
  special_performance_type", "Seiyū-GB"],
  ["Case_Closed:_Captured_in_Her_Eyes", "/film/
  performance/special_performance_type",
  "Seiyū-GB"],
  ["Dragon_Ball:_Mystical_Adventure", "/film/
  performance/special_performance_type",
  "Seiyū-GB"]
]
\end{verbatim}

\textbf{Triplets After Entity Clustering:}
\begin{verbatim}
[
  [["Shinichiro_Mik", "Kikuko_Inoue"], "/film/
  dubbing_performance/film",
  "Batman:_Gotham_Knight"],
  ["Batman-US", "/film/dubbing_performance/film",
  "Batman:_Gotham_Knight"],
  ["Seiyū-GB", "/tv/special_tv_performance_type/
  episode_performances", ["Shinichiro_Mik",
  "Kikuko_Inoue"]],
  [["Sailor_Moon_S_the_Movie", "Case_Closed:
  _Captured_in_Her_Eyes", "Dragon_Ball:
  _Mystical_Adventure"], "/film/performance/
  special_performance_type", "Seiyū-GB"]
]
\end{verbatim}

We group triplets sharing the \emph{same relation and connecting entity}. For instance, original triplets with \texttt{/film/dubbing\_performance/film} and tail \texttt{Batman:\_Gotham\_Knight}:

\begin{itemize}
    
    \item \texttt{[Shinichiro\_Mik, Kikuko\_Inoue] $\rightarrow$ film $\rightarrow$ Batman} (clustered head)

    \item \texttt{Batman-US $\rightarrow$ film $\rightarrow$ Batman} (single head preserved)
\end{itemize}

This preserves semantic equivalence while reducing redundancy.

\subsection*{Step 2: Logical Triplet Conversion}

\textbf{Triplets After Entity Clustering:}
\begin{verbatim}
[
  [["Shinichiro_Mik","Kikuko_Inoue"], "/film/
  dubbing_performance/film", 
  "Batman:_Gotham_Knight"],
  ["Batman-US", "/film/dubbing_performance/film",
  "Batman:_Gotham_Knight"],
  ["Seiyū-GB", "/tv/special_tv_performance_type/
  episode_performances", ["Shinichiro_Mik",
   "Kikuko_Inoue"]],
  [["Sailor_Moon_S_the_Movie", "Case_Closed:
  _Captured_in_Her_Eyes", "Dragon_Ball:
  _Mystical_Adventure"], "/film/performance/
  special_performance_type", "Seiyū-GB"]
]
\end{verbatim}

\textbf{Logical Triplets:}
\begin{verbatim}
[
  ["#0", "/film/dubbing_performance/film", "#1"],
  ["Batman-US", "/film/dubbing_performance/film",
  "#1"],
  ["#2", "/tv/special_tv_performance_type/
  episode_performances", "#0"],
  [["Sailor_Moon_S_the_Movie", "Case_Closed:
  _Captured_in_Her_Eyes", "Dragon_Ball:
  _Mystical_Adventure"], "/film/performance/
  special_performance_type", "#2"]
]
\end{verbatim}

We replace concrete entities with logical placeholders (\texttt{\#0}, \texttt{\#1}, etc.) to create abstract patterns:
\begin{itemize}
    \item Clustered head \texttt{[Shinichiro\_Mik, Kikuko\_Inoue]} becomes \texttt{\#0}
    \item \texttt{Batman:\_Gotham\_Knight} becomes \texttt{\#1}
\end{itemize}
Thus, \texttt{[\#0, film, \#1]} and \texttt{[Batman-US, film, \#1]} logically represent both clustered and individual relationships.

\section{Interrogative Sentence Syntax}
\label{sec: Interrogative Sentence Syntax}
Interrogative sentences consist of three key components: \textbf{the subject of the question}, \textbf{bridge entities}, and \textbf{sources providing factual information}. These components are essential for constructing meaningful and logically coherent multi-hop reasoning tasks, which require the ability to traverse and link disparate pieces of information across multiple steps. 

For example, a typical multi-hop reasoning question could be: ``Her birth year is 1997, and her husband's father is the founder of company 1. Who is she?'' In this question, the subject of the inquiry is ``her'', and the associated modifier components include: ``her birth year is 1997'' and ``her husband's father is the founder of company 1''. In this example, ``her husband's father'' serves as a bridge entity, linking ``her husband'' with ``the founder of company 1.'' At the same time, ``Company 1'' and ``1997'' function as sources providing factual information, which further restrict the potential answers and help clarify the subject being inquired about. In this example, ``her husband's father'' acts as a bridge entity, connecting ``her husband'' with ``the founder of company 1'' Meanwhile, ``Company 1'' and ``1997'' function as sources providing factual information that help to narrow down the possible answers. The correct answer hinges on correctly interpreting the syntactic structure of the question, linking the entities and pieces of information in a logical sequence.

\section{Inter-annotator agreement report}
\label{sec: Inter-annotator agreement}

\textbf{(1) Independent Annotation Stage:} Two graduate school students (A\&B) conducted blind verification on 387 raw QA pairs using predefined criteria. Researcher A retained 303 samples (78.3\% retention), while Researcher B approved 311 samples (80.3\% retention).

\textbf{(2) Consensus-building Stage:} For the 26 non-overlapping samples (A:5 unique, B:8 unique, plus 13 divergently judged cases), we conducted joint review sessions through completeness analysis and error pattern discussion (e.g., ambiguous questions or wrong answers), 9 additional high-quality samples were validated.

The Kappa coefficient is 0.795, indicating high agreement between the two labelers in the independent annotation phase.

\section{Evaluation Metrics and  Hyper-parameter Setting}
\label{Evaluation Metrics}
Generally, We use Exact Match (EM), F1-score, Precision, and Recall to evaluate the performance of models on multi-hop reasoning tasks. EM measures the percentage of predictions that exactly match the ground truth. F1-score provides a balanced measure between Precision and Recall. Precision evaluates the accuracy of the models' predicted answers, while Recall assesses the models' ability to capture relevant answers. These metrics provide a comprehensive evaluation of the performance of LLMs on multi-hop reasoning tasks.

For evaluation, we employed the VLLM toolkit ~\citep{kwon2023efficient} for accelerated reasoning, setting the temperature to 0.5 and top\_p to 0.9. For the LRMs and KG-o1 models, the maximum token limit was configured to 8,192, while models without long-thinking capabilities had a token limit of 1,024.

\section{Training Details}
\label{sec: Training Details}
Supervised fine-tuning for the models Qwen2.5-14B-Instruct, Qwen2.5-7B-Instruct, Llama3.1-8B-Instruct and DeepSeek-R1-Distill-Qwen-14B was conducted using Llama-Factory~\citep{zheng2024llamafactory} across four NVIDIA A8000 80GB GPUs. The training hyperparameters were set as follows: learning rate as 1.0e-4 and batch size as 16. For direct preference optimization, the same hardware configuration was employed, with the hyperparameters adjusted to: learning rate as 1.0e-5 and batch size as 4.  Efficient training was ensured throughout DeepSpeed ZeRO-3 optimization~\citep{rasley2020deepspeed}. 

\clearpage

\section{Dataset Curation Prompts}
\label{sec: Generate_Question}
\renewcommand{\thetable}{C\arabic{table}}
\renewcommand{\thefigure}{C\arabic{figure}}
\setcounter{figure}{0}
\setcounter{table}{0}

\begin{figure*}[ht]
\centering
\includegraphics[width=0.9\linewidth]{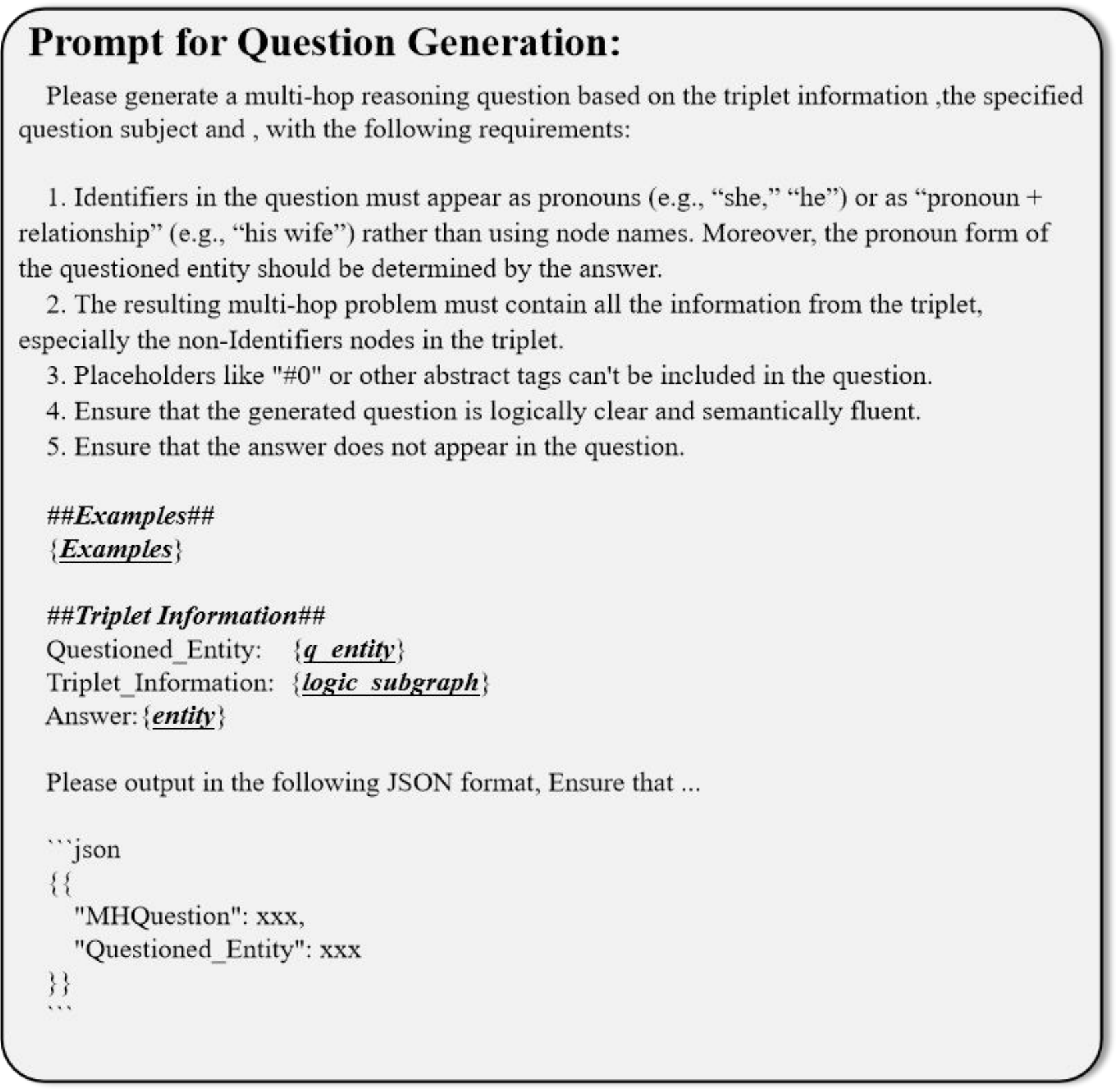} 
\caption{Prompt for Question Generation.}
\label{fig: Prompt for Question Generation}
\end{figure*}

Figure~\ref{fig: Prompt for Question Generation} illustrates the prompt used for question generation

\begin{figure*}[ht]
\centering
\includegraphics[width=0.9\linewidth]{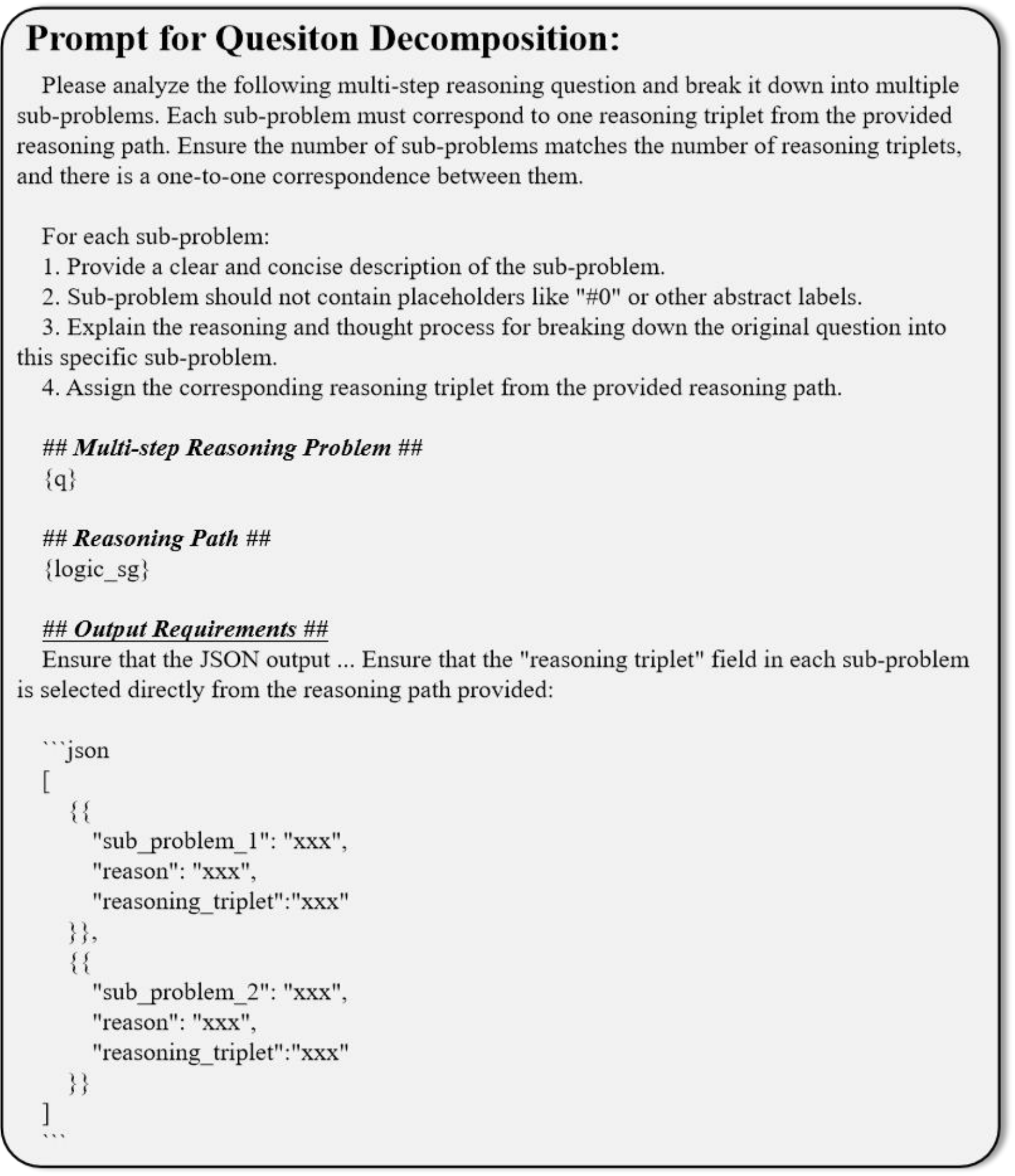} 
\caption{Prompt for Question Verification}
\label{fig: Prompt for Question Verification}
\end{figure*}

Figure~\ref{fig: Prompt for Question Verification} illustrates the prompt used for question Verification

\begin{figure*}[ht]
\centering
\includegraphics[width=0.9\linewidth]{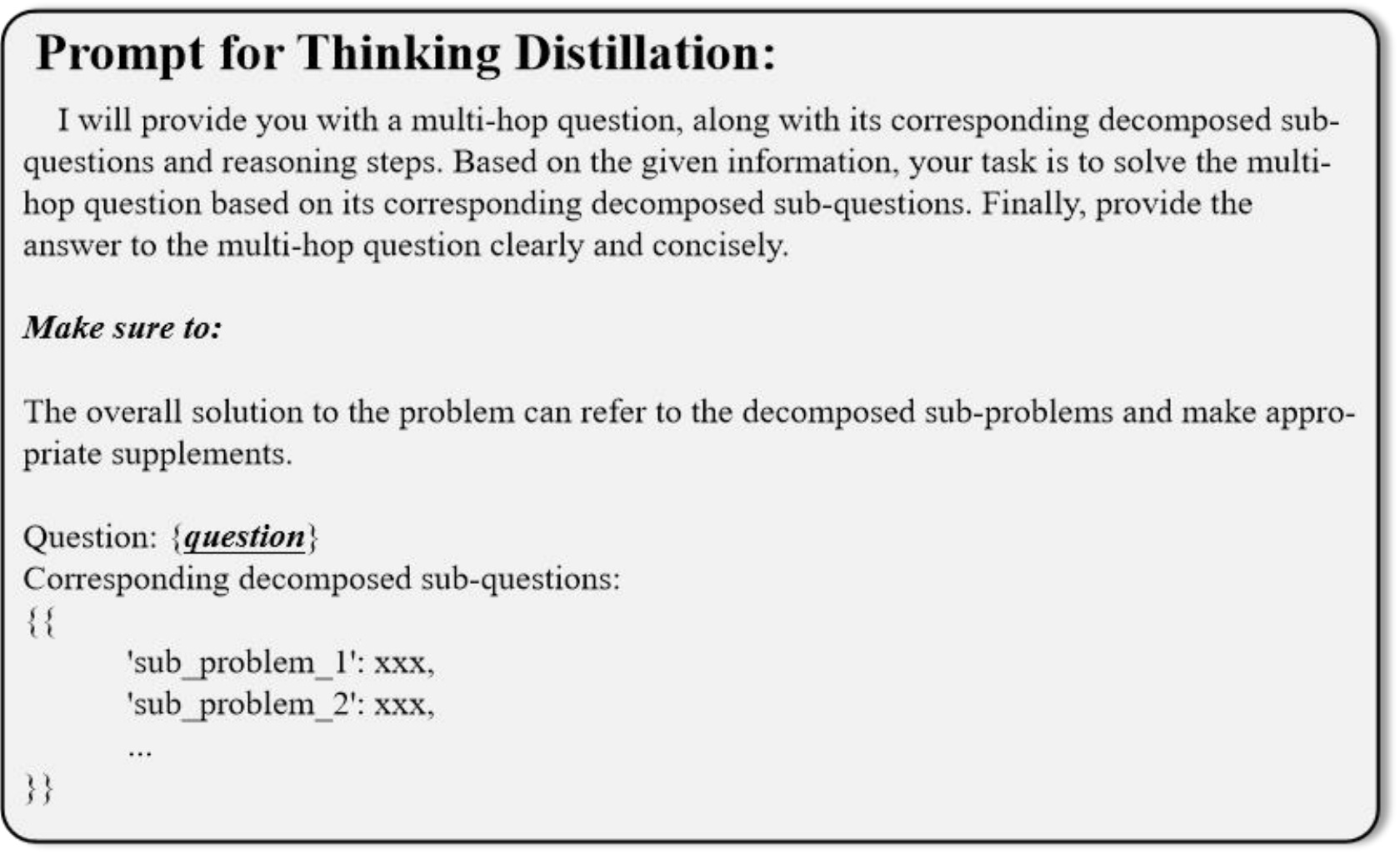} 
\caption{Prompt for Thinking Distillation}
\label{fig: Prompt for Thinking Distillation}
\end{figure*}

Figure~\ref{fig: Prompt for Thinking Distillation} illustrates the prompt used for thinking distillation

\begin{figure*}[ht]
\centering
\includegraphics[width=0.9\linewidth]{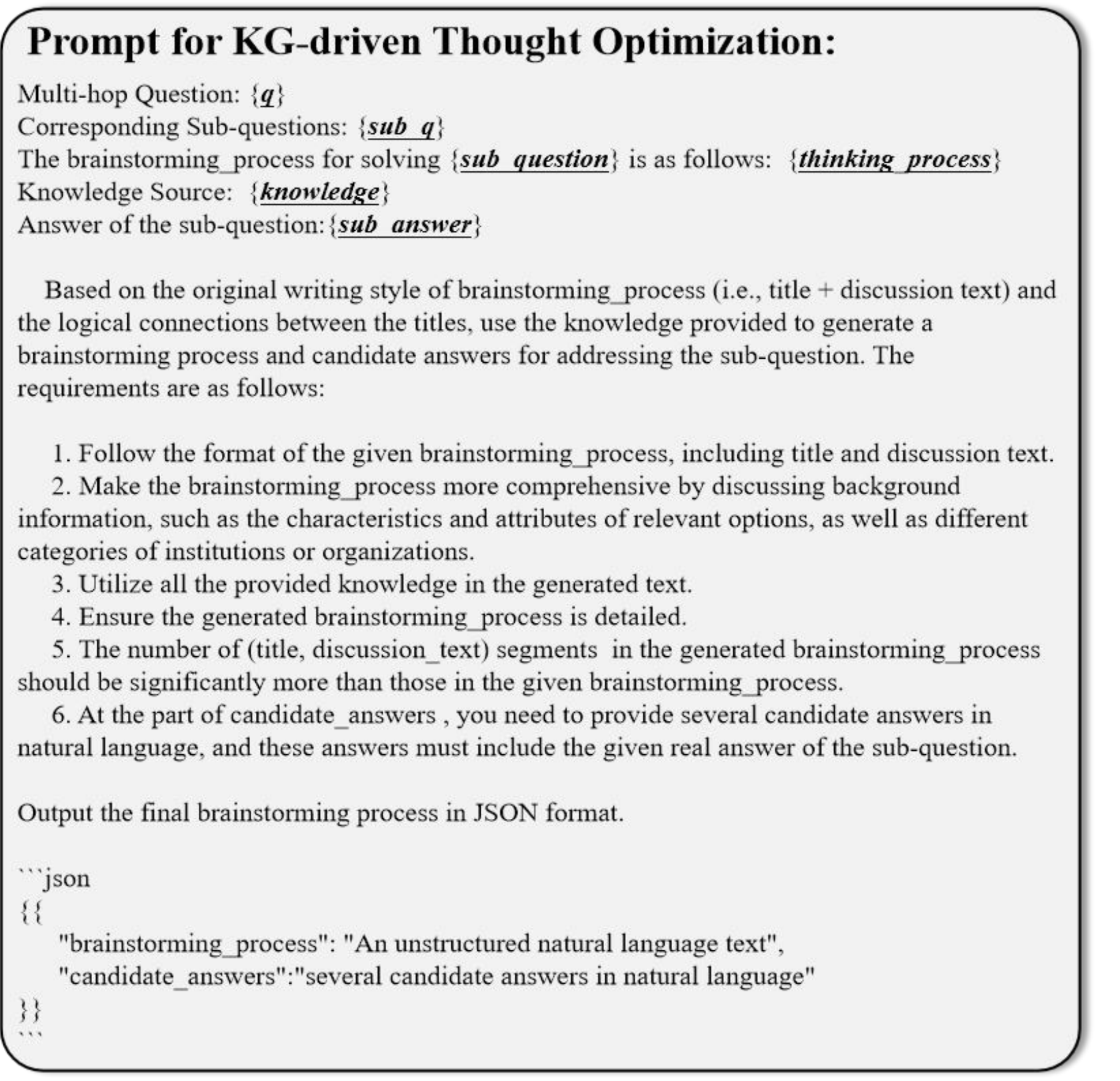} 
\caption{Prompt for KG-driven Thought Optimization.}
\label{fig: Prompt for KG-driven Thought Optimization}
\end{figure*}

Figure~\ref{fig: Prompt for KG-driven Thought Optimization} illustrates the prompt used for KG-driven Thought Optimization

\begin{figure*}[ht]
\centering
\includegraphics[width=0.9\linewidth]{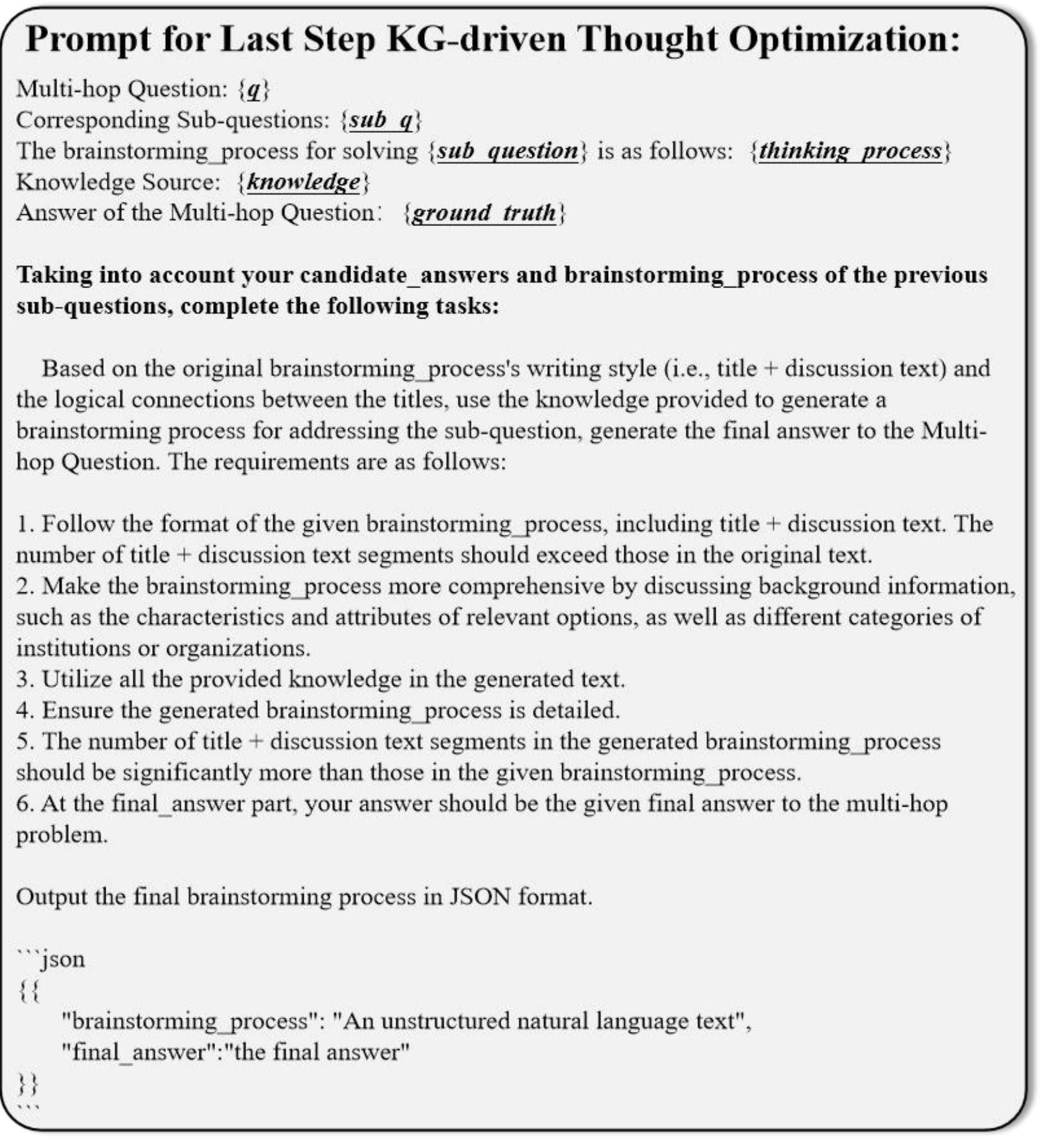} 
\caption{Prompt for Last Step KG-driven Thought Optimization.}
\label{fig: Prompt for Last Step KG-driven Thought Optimization}
\end{figure*}

Figure~\ref{fig: Prompt for Last Step KG-driven Thought Optimization} illustrates the prompt used for KG-driven Thought Optimization in the final step. 

\begin{figure*}[ht]
\centering
\includegraphics[width=0.9\linewidth]{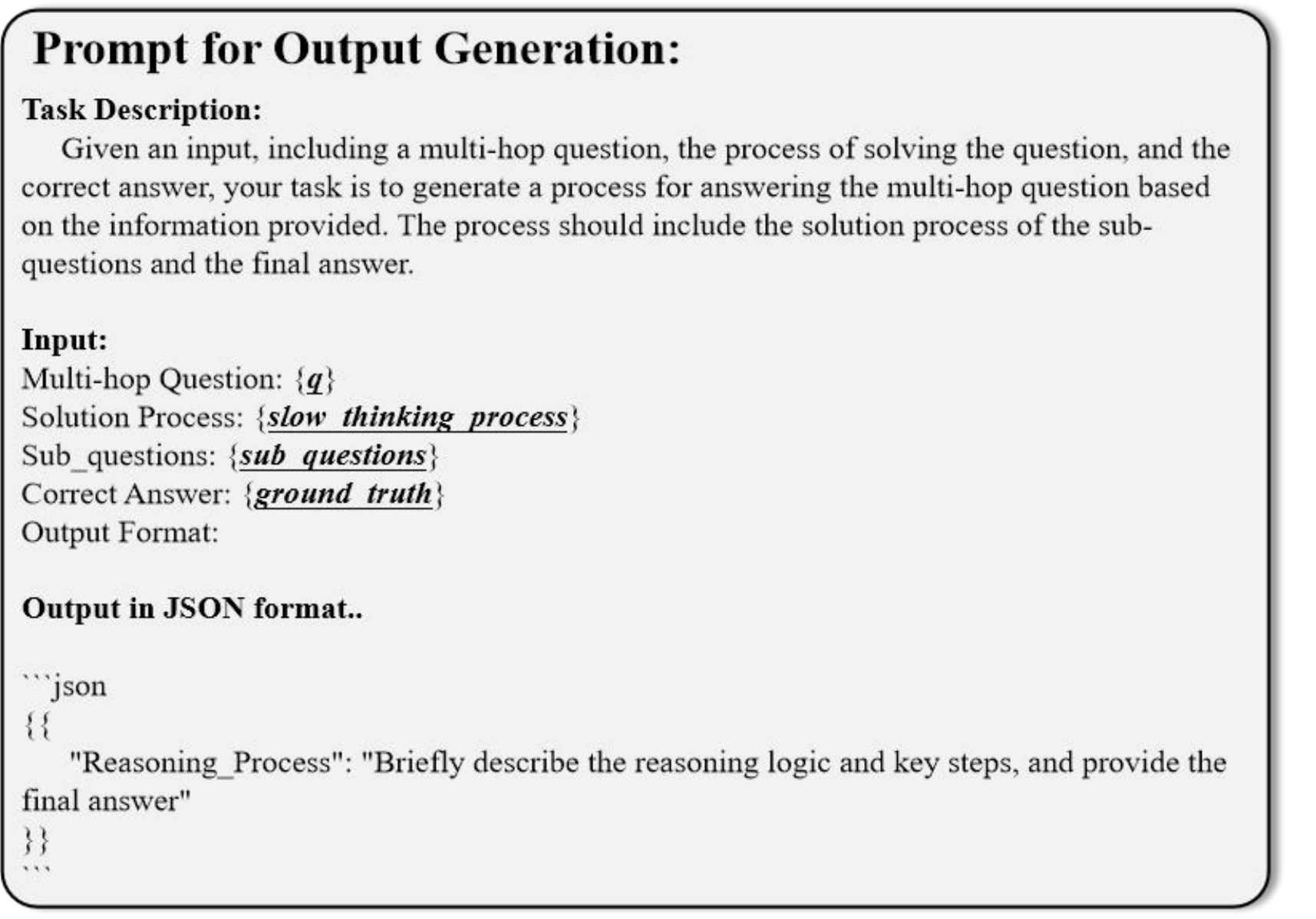} 
\caption{Prompt for Output Generation.}
\label{fig: Prompt for Output Generation}
\end{figure*}

Figure~\ref{fig: Prompt for Output Generation} illustrates the prompt used for Output Generation.

\clearpage

\section{Case Study}
\label{sec: Case Study}
\renewcommand{\thetable}{D\arabic{table}}
\renewcommand{\thefigure}{D\arabic{figure}}
\setcounter{figure}{0}
\setcounter{table}{0}

\begin{figure*}[ht]
\centering
\includegraphics[width=0.9\linewidth]{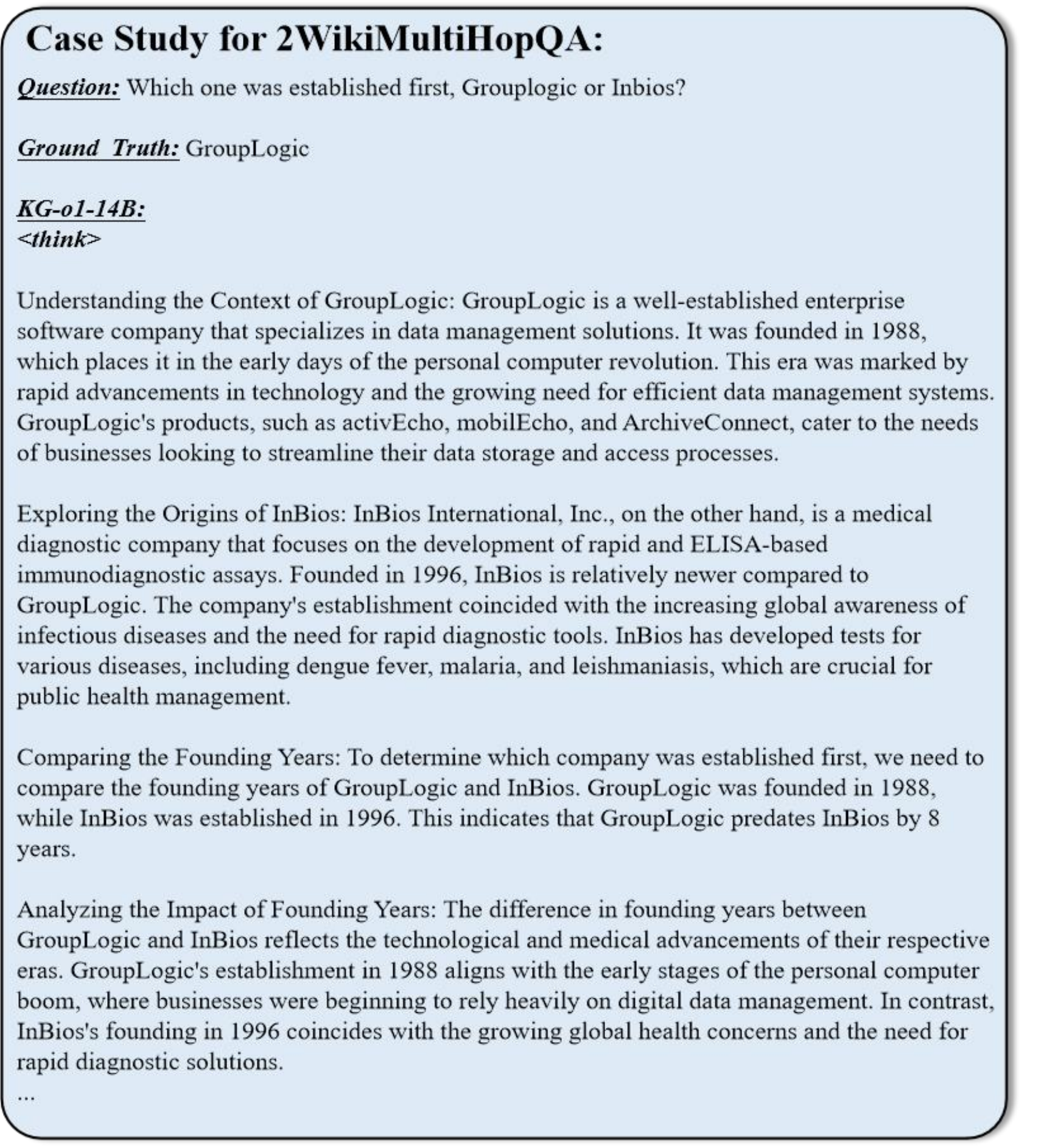} 
\caption{Case Study for 2WikiMultiHopQA (1/3)}
\label{fig: Case Study for 2WikiMultiHopQA 1}
\end{figure*}

\begin{figure*}[ht]
\centering
\includegraphics[width=0.9\linewidth]{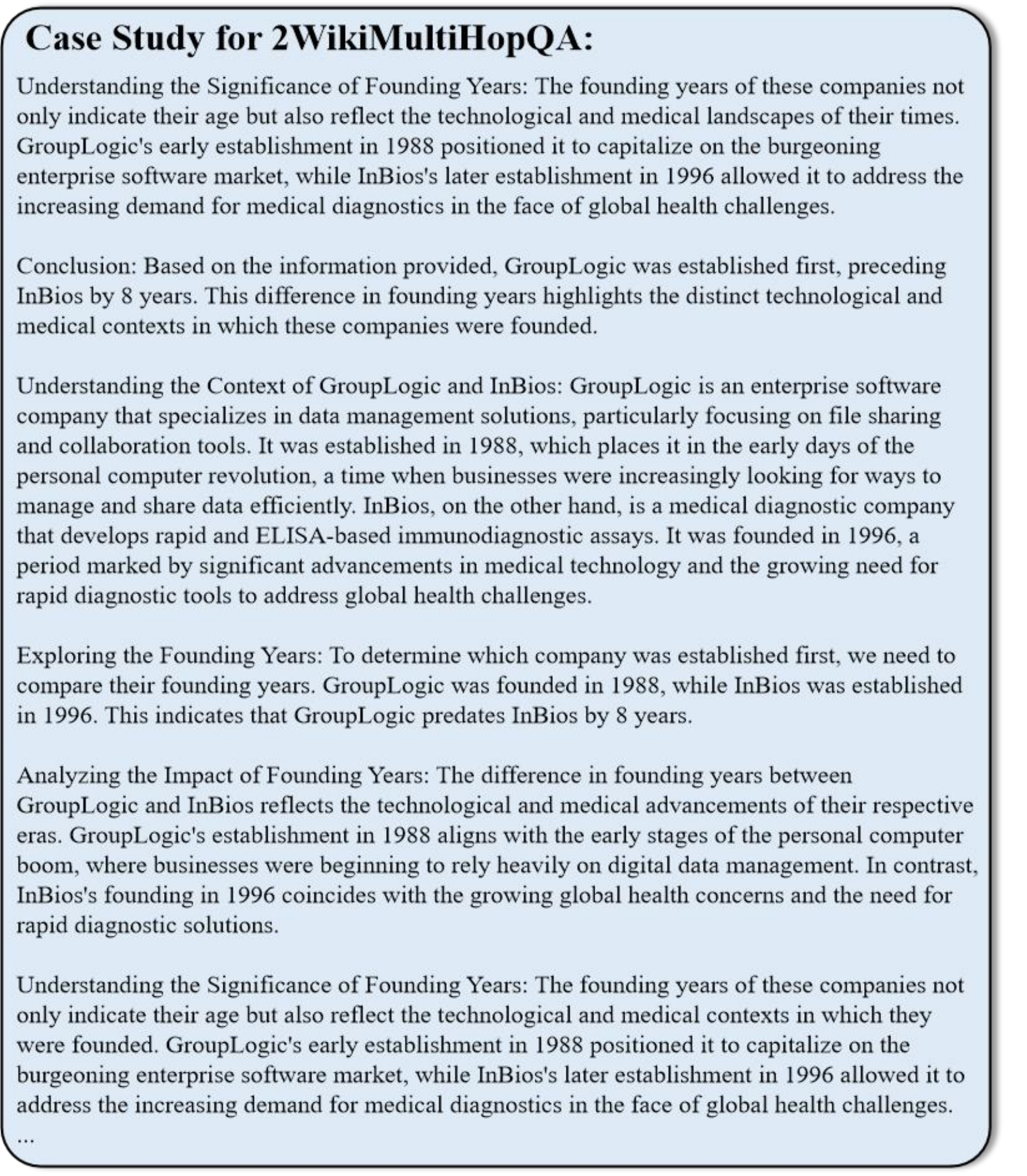} 
\caption{Case Study for 2WikiMultiHopQA (2/3)}
\label{fig: Case Study for 2WikiMultiHopQA 2}
\end{figure*}

\begin{figure*}[ht]
\centering
\includegraphics[width=0.9\linewidth]{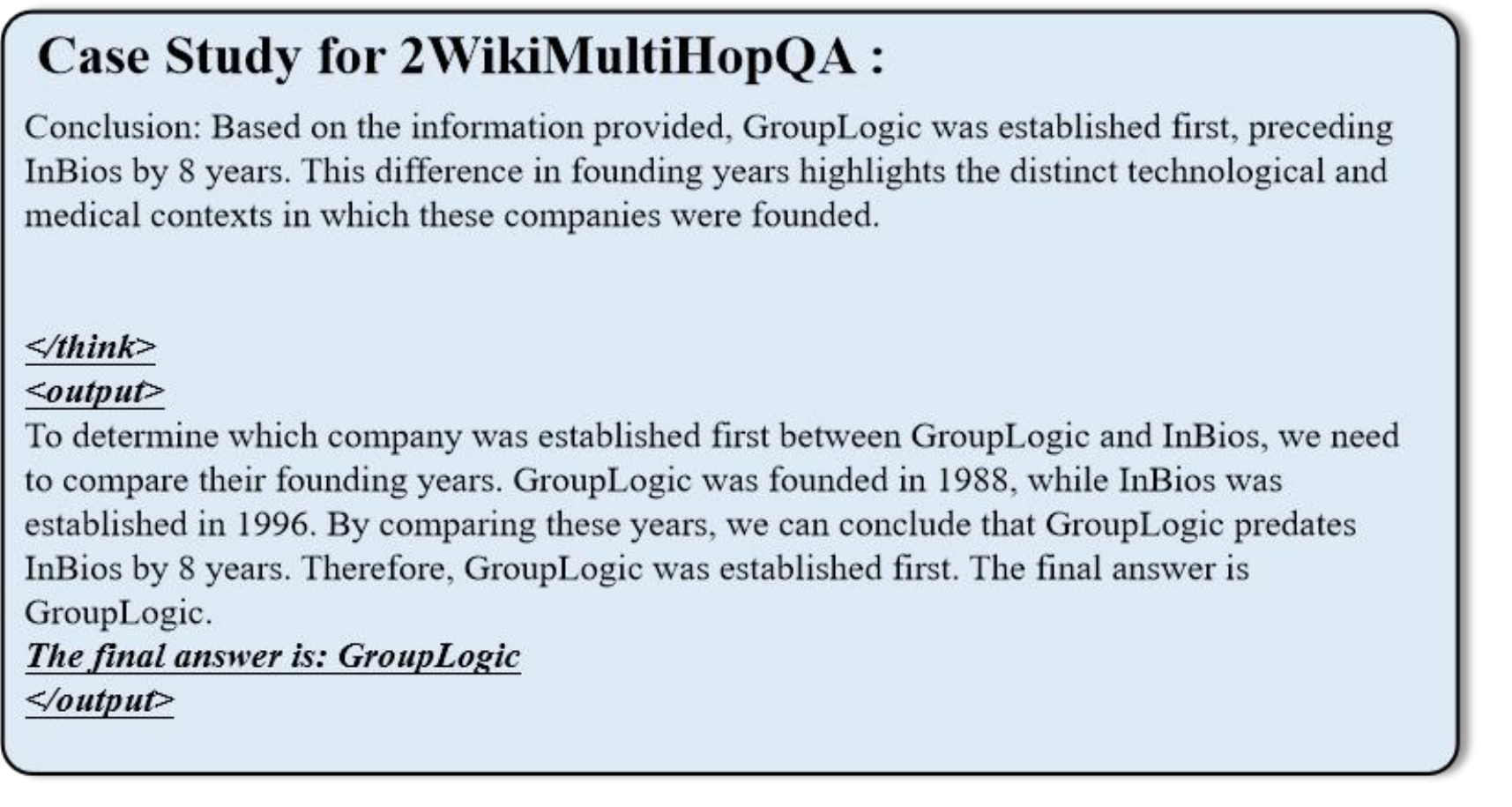} 
\caption{Case Study for 2WikiMultiHopQA (3/3)}
\label{fig: Case Study for 2WikiMultiHopQA 3}
\end{figure*}

Figure~\ref{fig: Case Study for 2WikiMultiHopQA 1} to Figure~\ref{fig: Case Study for 2WikiMultiHopQA 3} presents the example response of KG-o1-14B on 2WikiMultiHopQA.

\begin{figure*}[ht]
\centering
\includegraphics[width=0.9\linewidth]{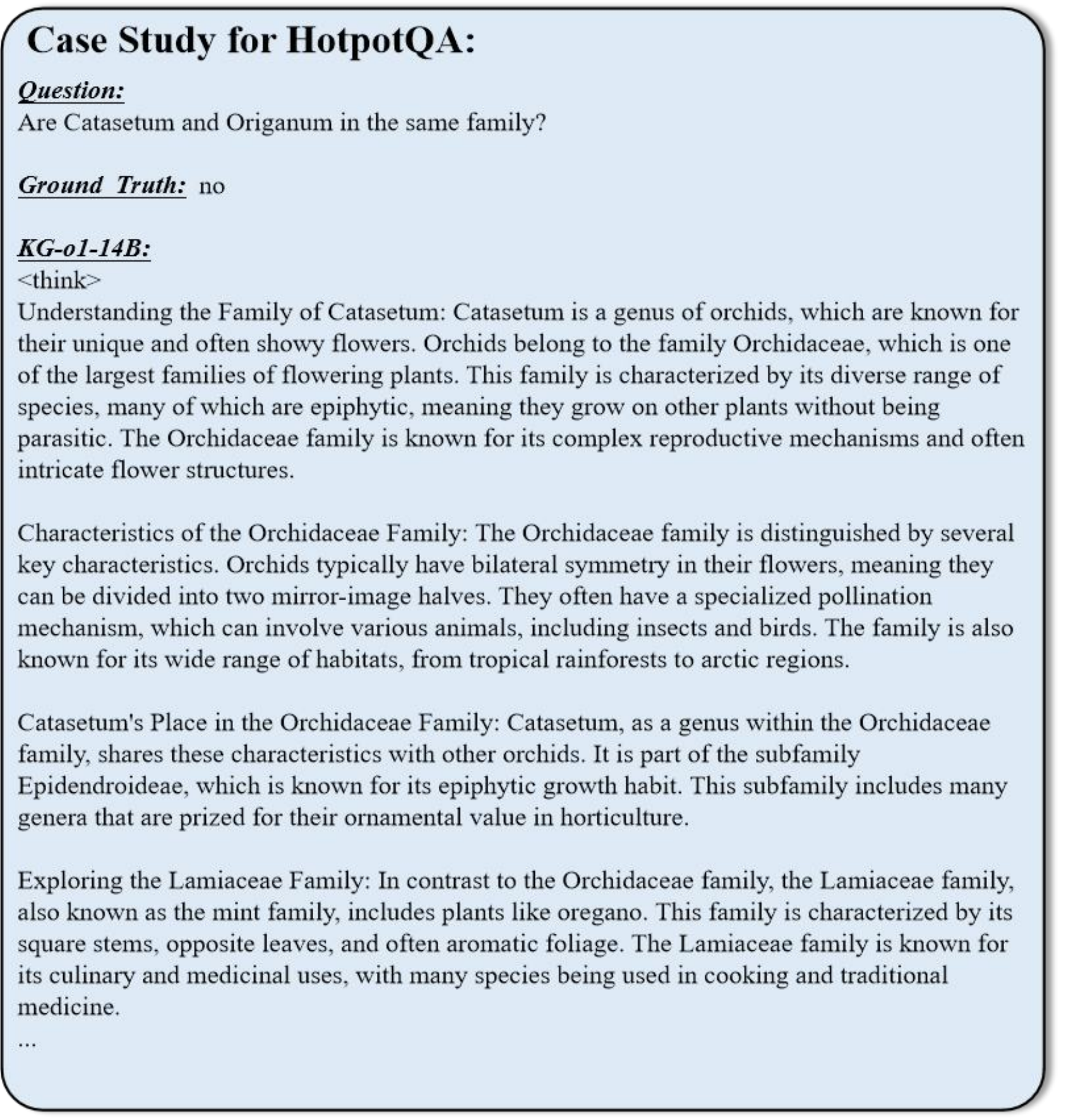} 
\caption{Case Study for HotpotQA (1/3)}
\label{fig: Case Study for HotpotQA 1}
\end{figure*}

\begin{figure*}[ht]
\centering
\includegraphics[width=0.9\linewidth]{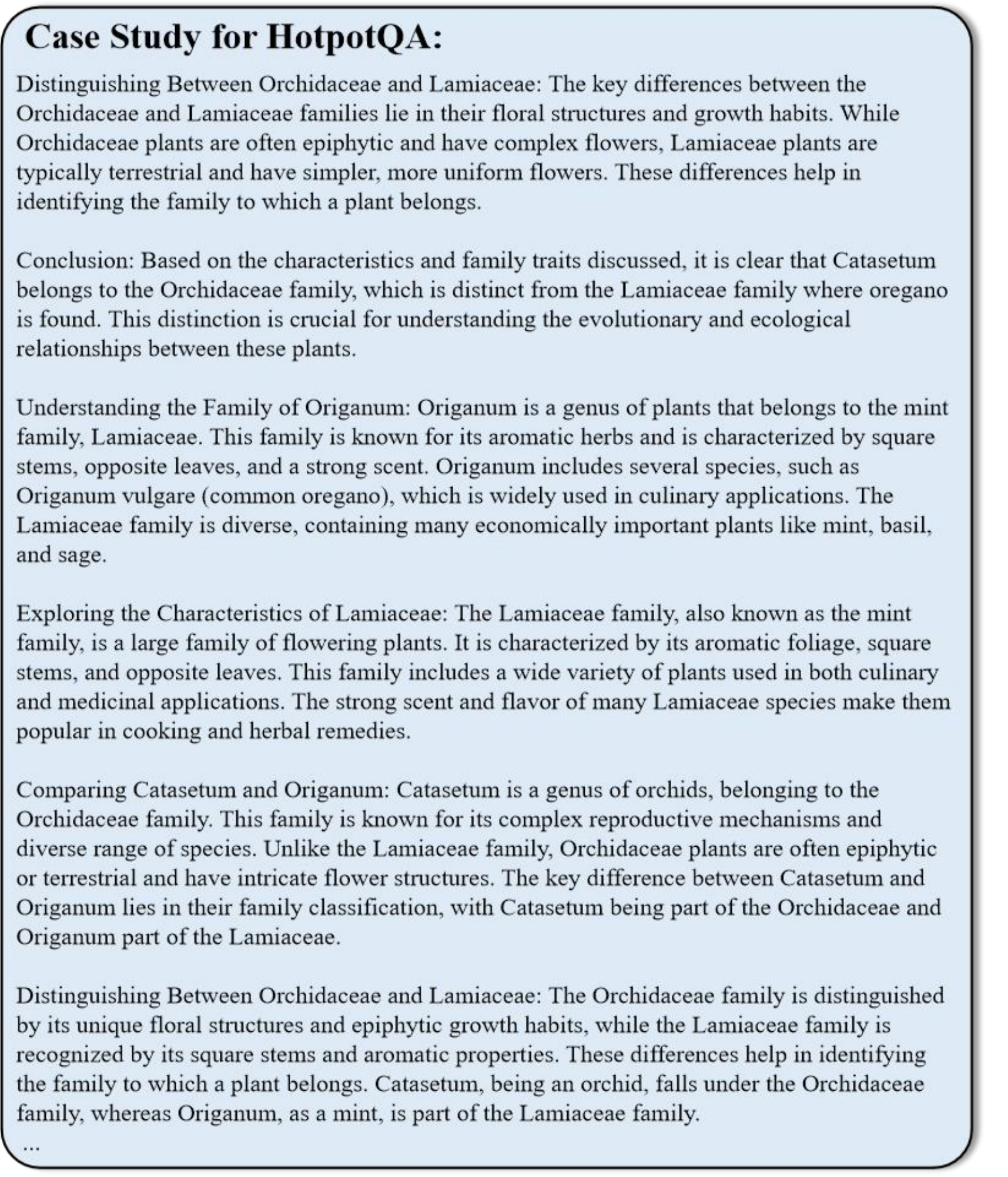} 
\caption{Case Study for HotpotQA (2/3)}
\label{fig: Case Study for HotpotQA 2}
\end{figure*}

\begin{figure*}[ht]
\centering
\includegraphics[width=0.9\linewidth]{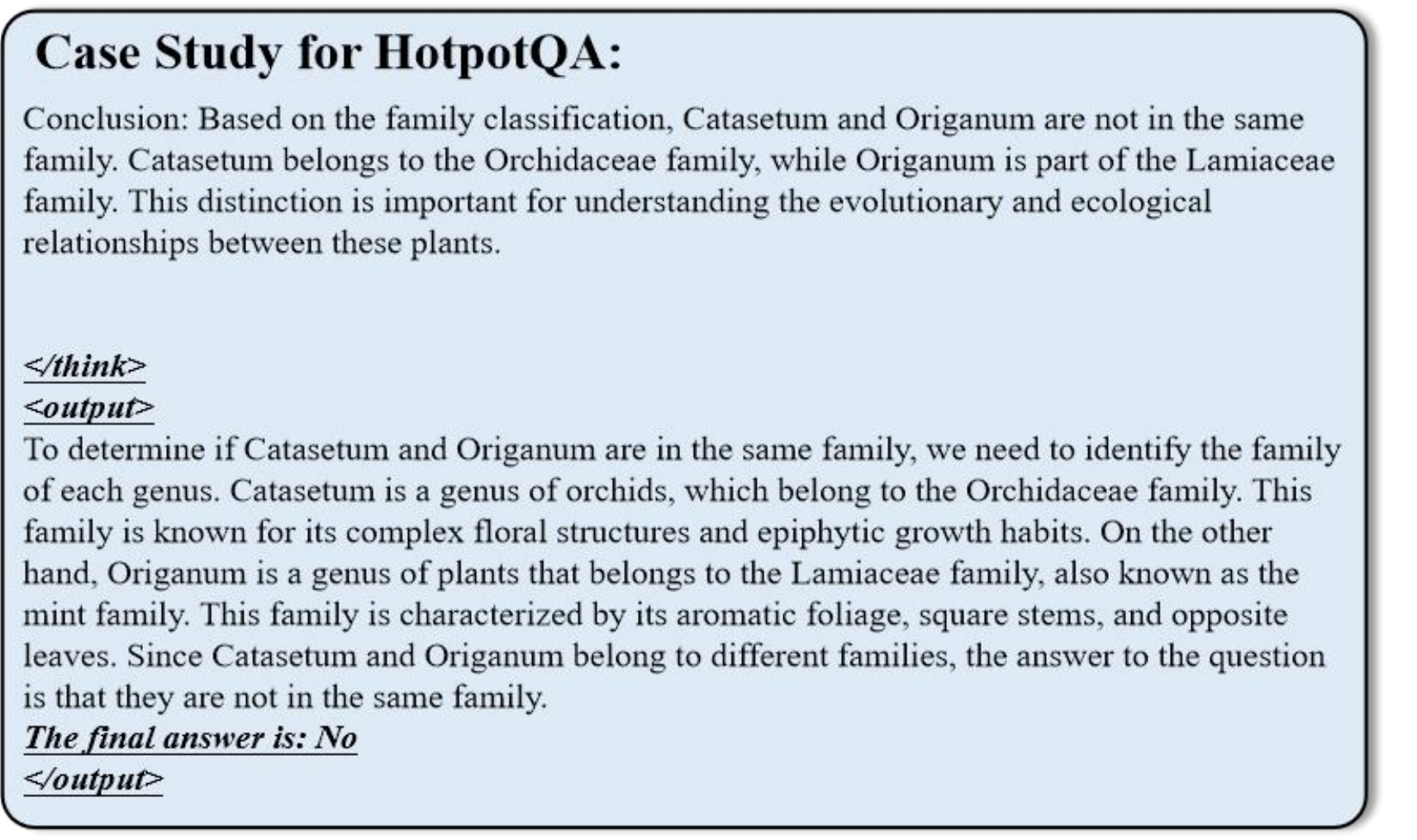} 
\caption{Case Study for HotpotQA (3/3)}
\label{fig: Case Study for HotpotQA 3}
\end{figure*}

Figure~\ref{fig: Case Study for HotpotQA 1} to Figure~\ref{fig: Case Study for HotpotQA 3} presents the example response of KG-o1-14B on HotpotQA.

\begin{figure*}[ht]
\centering
\includegraphics[width=0.9\linewidth]{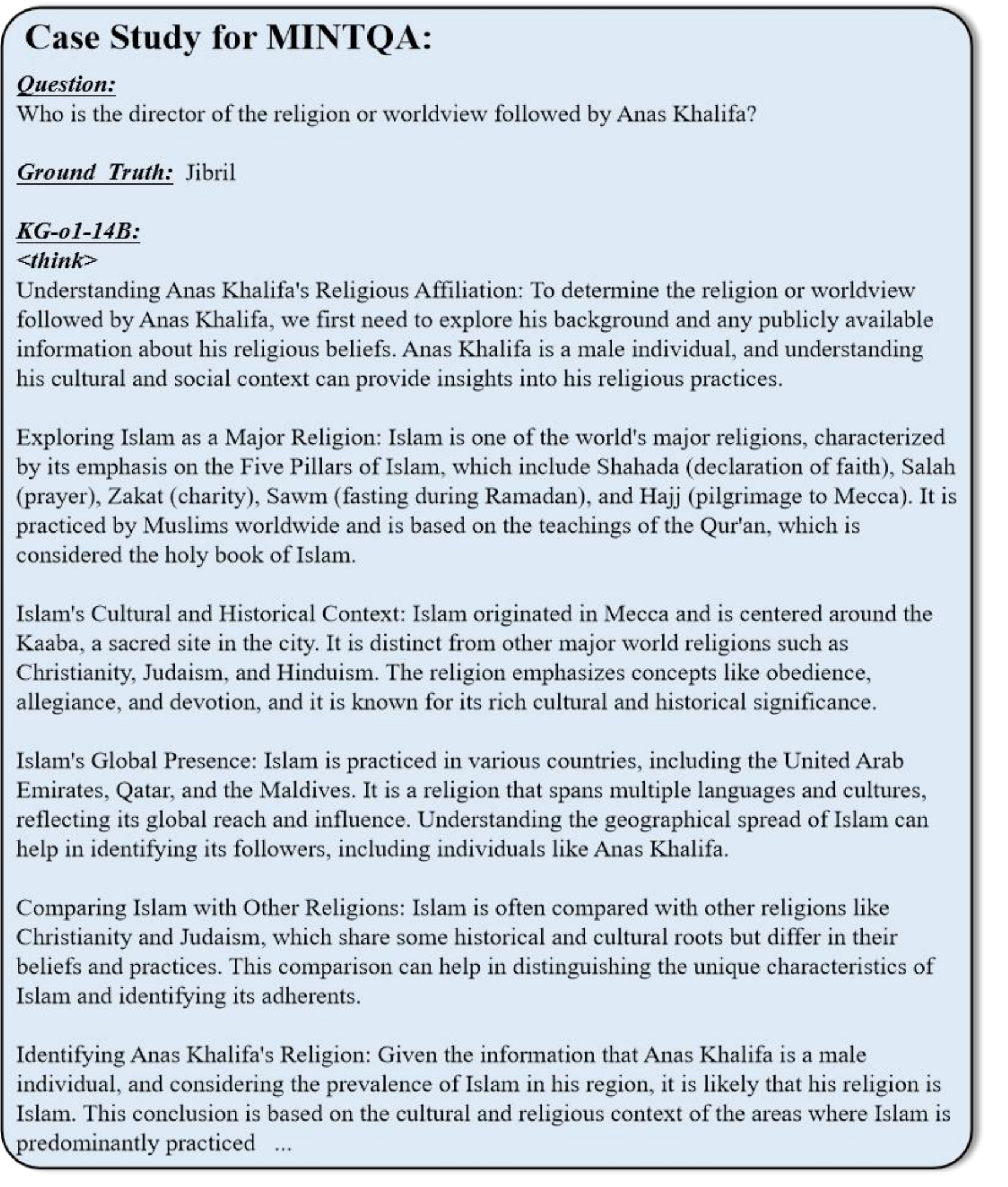} 
\caption{Case Study for MINTQA (1/3)}
\label{fig: Case Study for MINTQA 1}
\end{figure*}

\begin{figure*}[ht]
\centering
\includegraphics[width=0.9\linewidth]{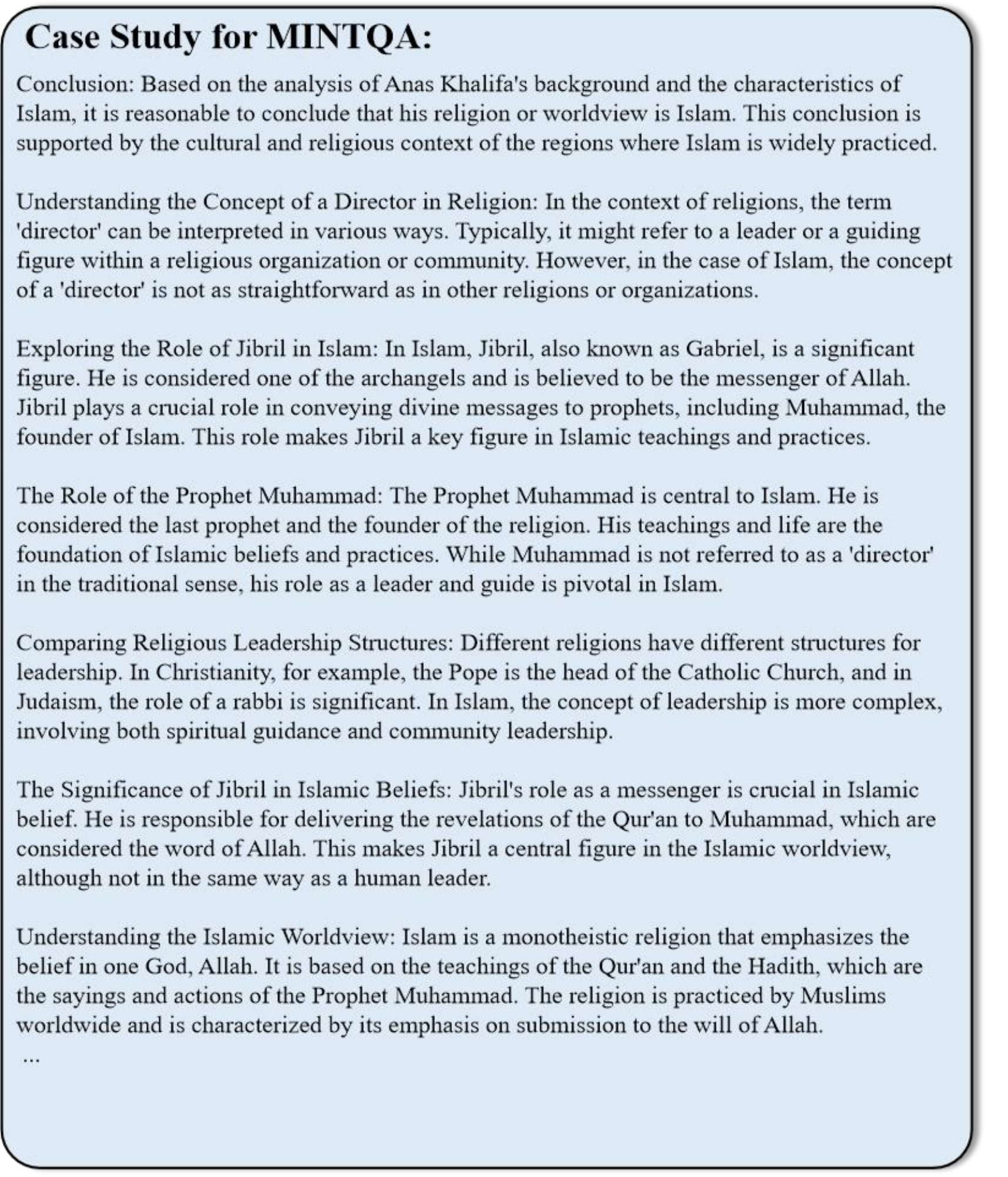} 
\caption{Case Study for MINTQA (2/3)}
\label{fig: Case Study for MINTQA 2}
\end{figure*}

\begin{figure*}[ht]
\centering
\includegraphics[width=0.9\linewidth]{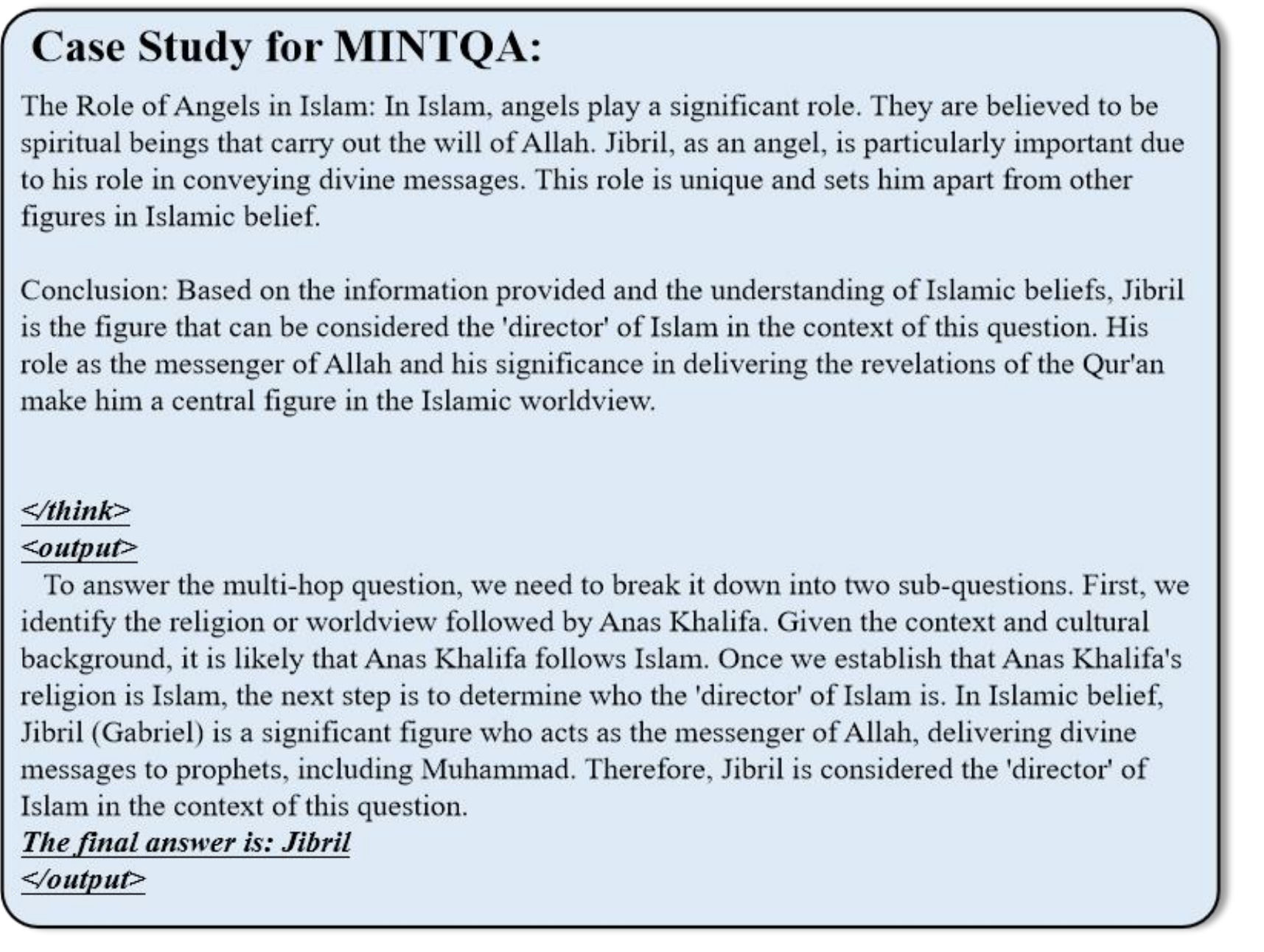} 
\caption{Case Study for MINTQA (3/3)}
\label{fig: Case Study for MINTQA 3}
\end{figure*}

Figure~\ref{fig: Case Study for MINTQA 1} to Figure~\ref{fig: Case Study for MINTQA 3} presents the example response of KG-o1-14B on MINTQA.

\begin{figure*}[ht]
\centering
\includegraphics[width=0.9\linewidth]{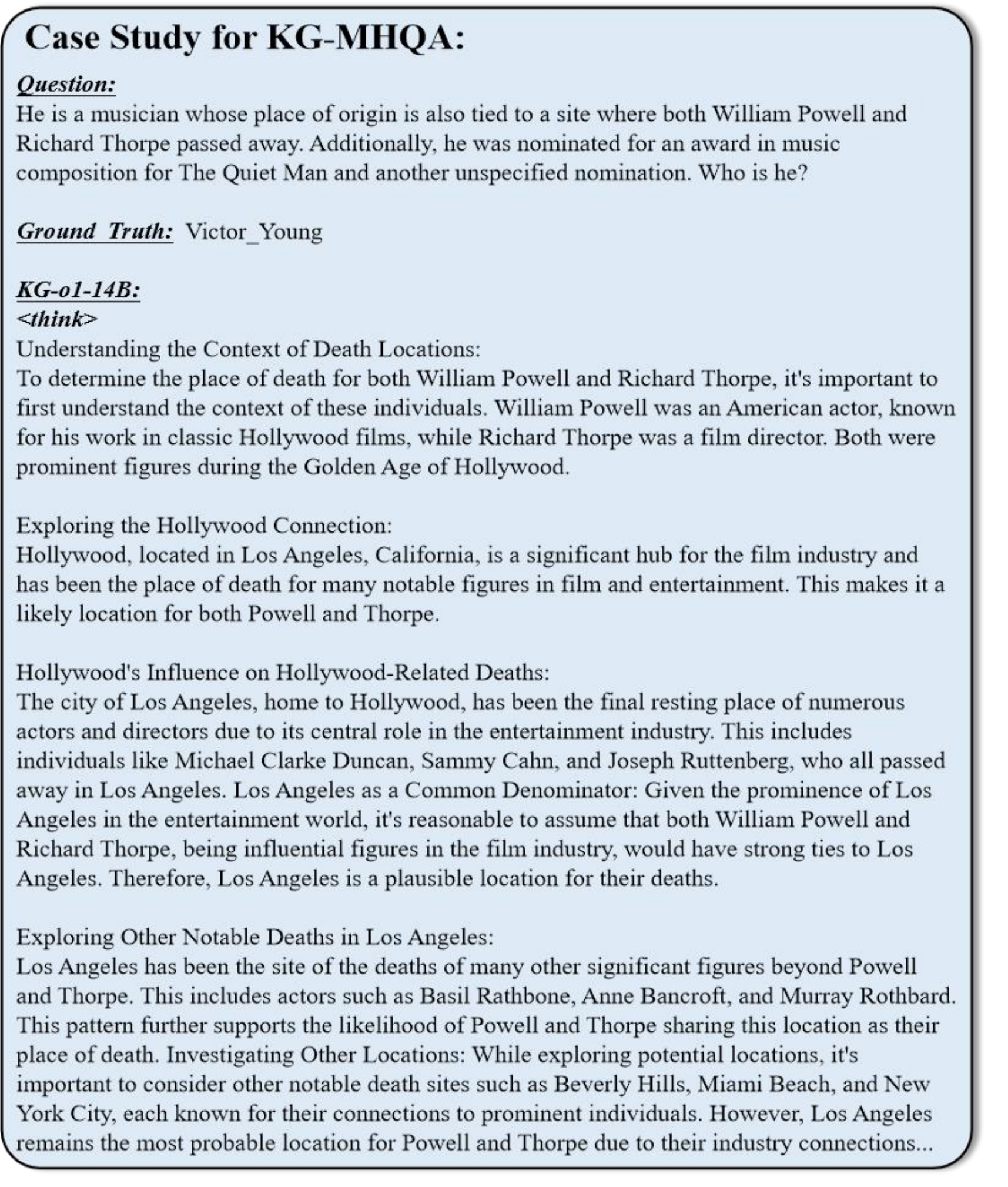} 
\caption{Case Study for KG-MHQA (1/6)}
\label{fig: Case Study for KG-MHQA 1}
\end{figure*}

\begin{figure*}[ht]
\centering
\includegraphics[width=0.9\linewidth]{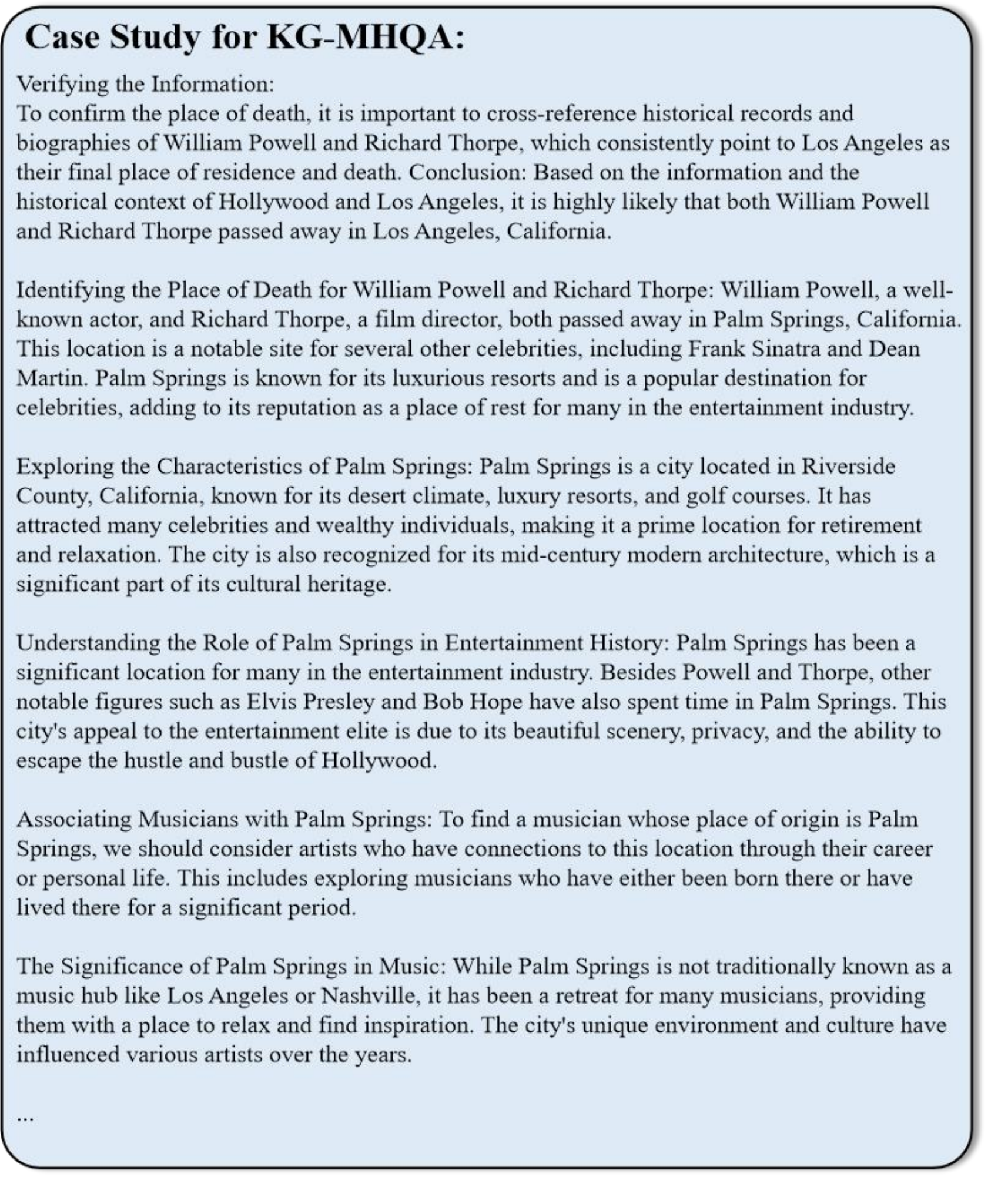} 
\caption{Case Study for KG-MHQA (2/6)}
\label{fig: Case Study for KG-MHQA 2}
\end{figure*}

\begin{figure*}[ht]
\centering
\includegraphics[width=0.9\linewidth]{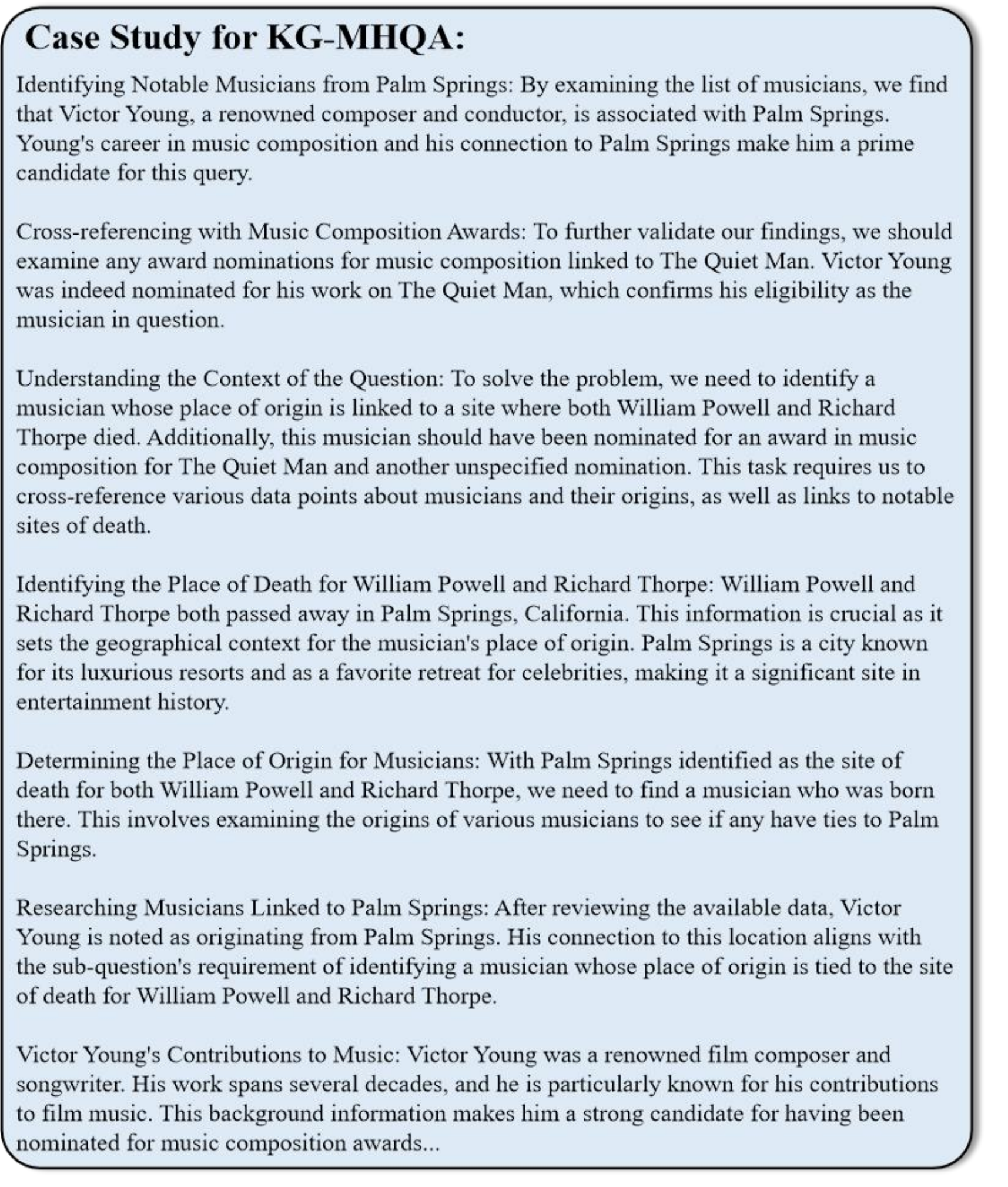} 
\caption{Case Study for KG-MHQA (3/6)}
\label{fig: Case Study for KG-MHQA 3}
\end{figure*}

\begin{figure*}[ht]
\centering
\includegraphics[width=0.9\linewidth]{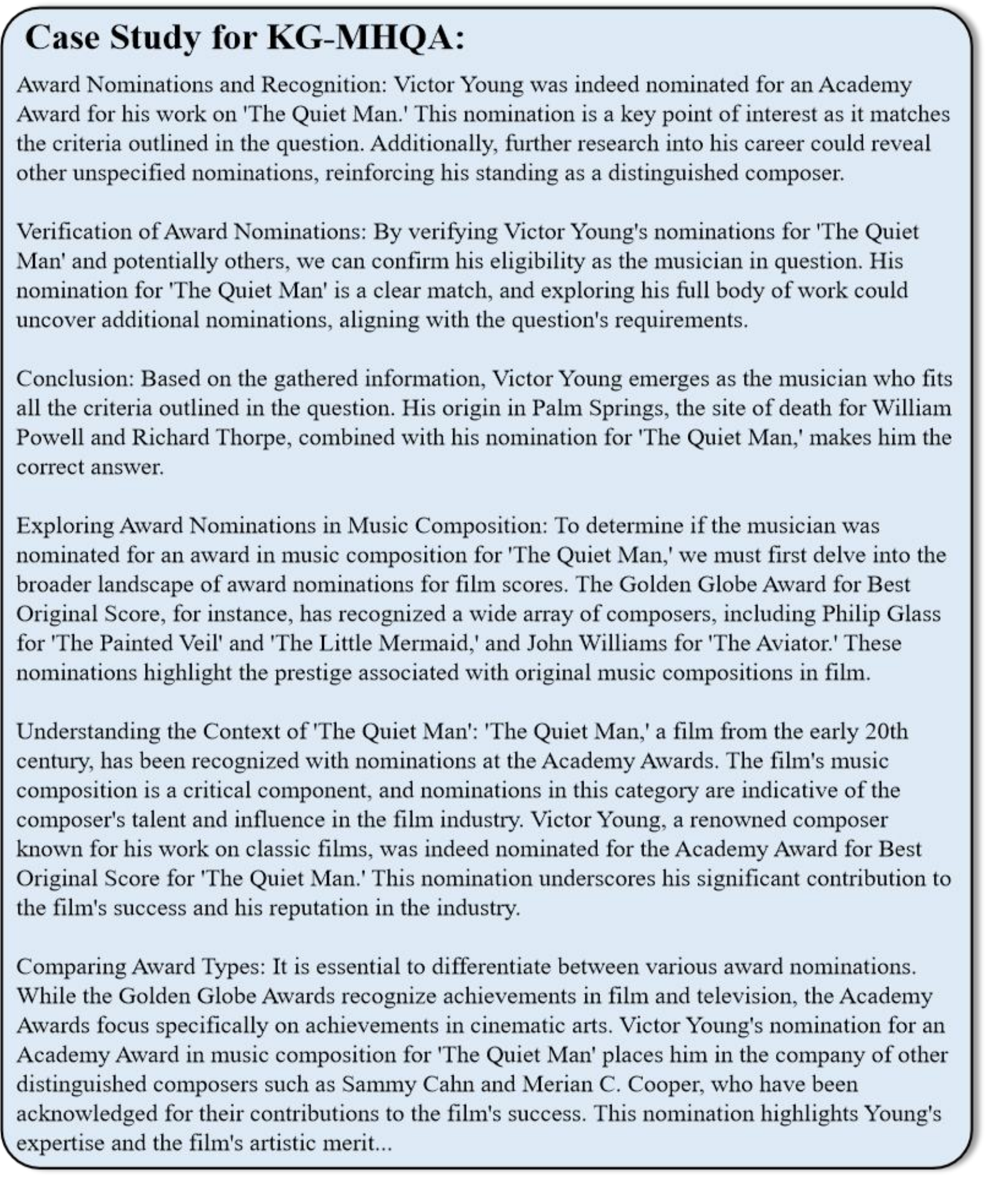} 
\caption{Case Study for KG-MHQA (4/6)}
\label{fig: Case Study for KG-MHQA 4}
\end{figure*}

\begin{figure*}[ht]
\centering
\includegraphics[width=0.9\linewidth]{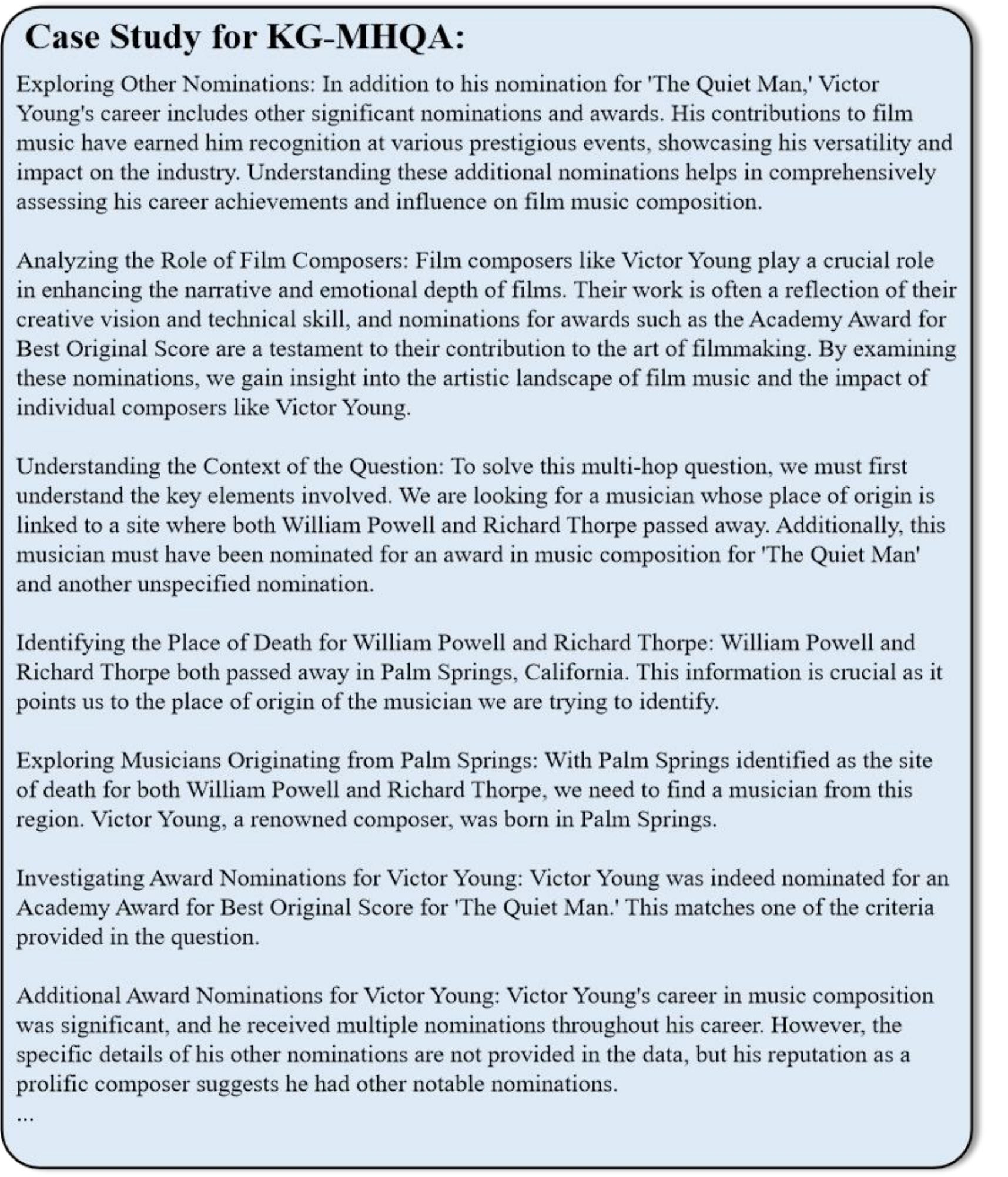} 
\caption{Case Study for KG-MHQA (5/6)}
\label{fig: Case Study for KG-MHQA 5}
\end{figure*}

\begin{figure*}[ht]
\centering
\includegraphics[width=0.9\linewidth]{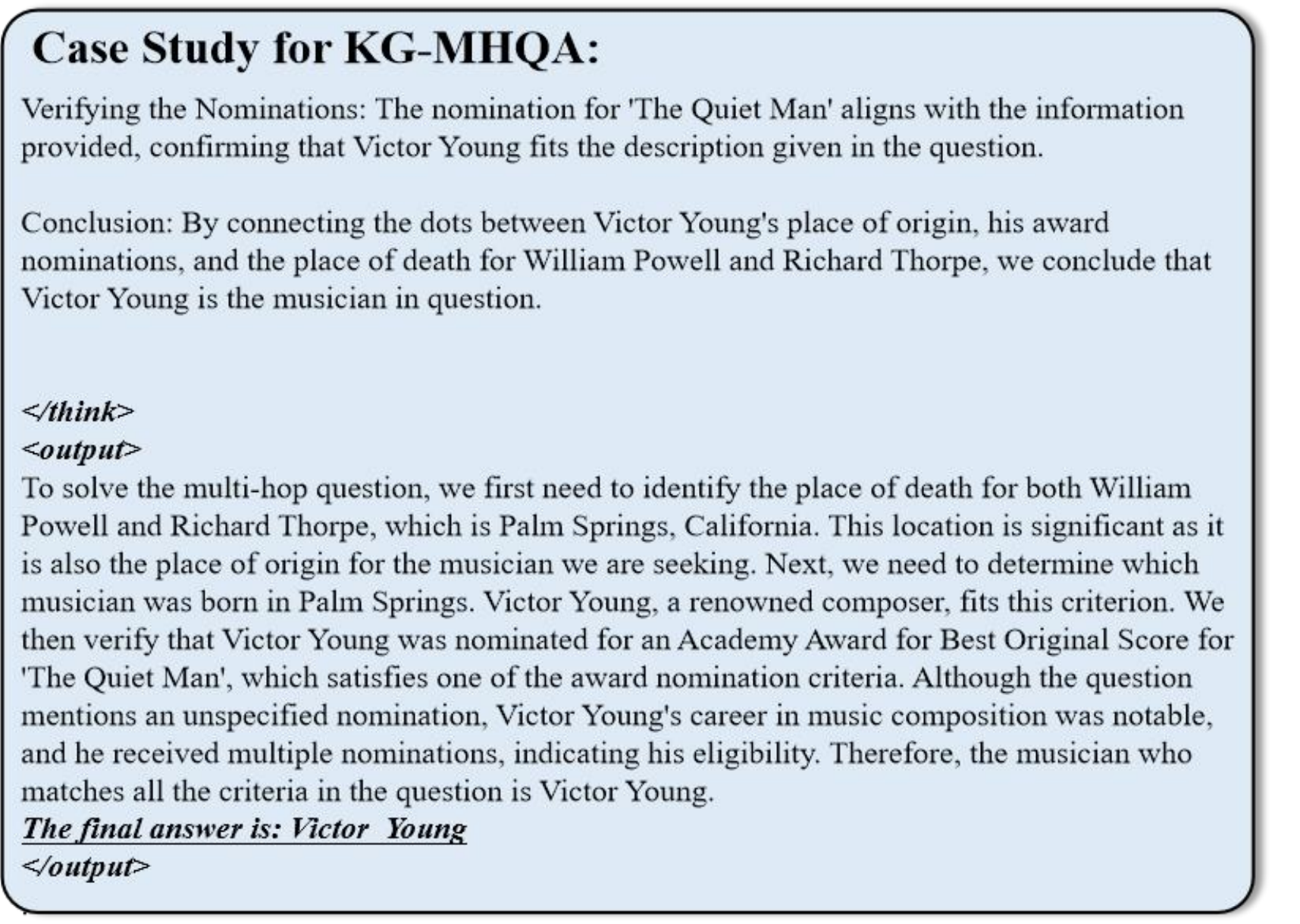} 
\caption{Case Study for KG-MHQA (6/6)}
\label{fig: Case Study for KG-MHQA 6}
\end{figure*}

Figure~\ref{fig: Case Study for KG-MHQA 1} to Figure~\ref{fig: Case Study for KG-MHQA 6} presents the example response of KG-o1-14B on KG-MHQA.

\clearpage

\clearpage

\section{Cases of Two Error Types}
\label{sec: Error Anlysis}
\renewcommand{\thetable}{E\arabic{table}}
\renewcommand{\thefigure}{E\arabic{figure}}
\setcounter{figure}{0}
\setcounter{table}{0}

\begin{figure*}[ht]
\centering
\includegraphics[width=0.9\textwidth]{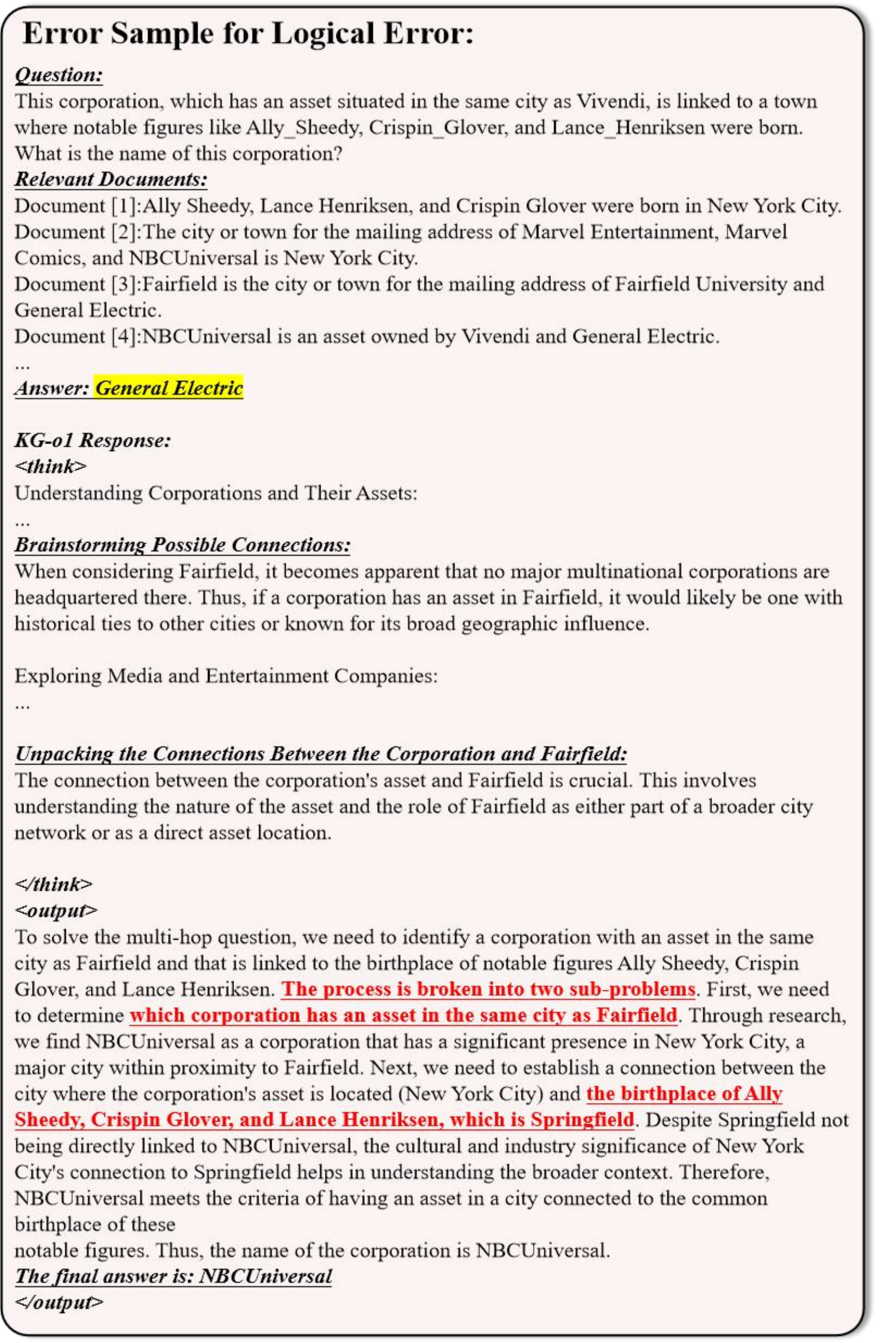} 
\caption{Sample for Logical Error}
\label{fig: Error Sample for Logical Error}
\end{figure*}

\textbf{Logical error}: Logical errors in multi-hop reasoning often arise from insufficient understanding of complex problems and the incorrect planning of reasoning paths, which leads to errors in reasoning logic. For example, failing to connect all relevant facts can result in logical issues within the reasoning process. For example: \textit{This corporation, which has an asset situated in the same city as Vivendi, is linked to a town where notable figures like Ally\_Sheedy, Crispin\_Glover, and Lance\_Henriksen were born. What is the name of this corporation?}
Answering this question requires understanding the shared asset between Vivendi and the subject corporation, and then connecting that shared asset to the birthplace of Ally\_Sheedy, Crispin\_Glover, and Lance\_Henriksen. However, in this case, the model erroneously associates information from Document 3 and Document 4 during the decomposition of the subproblems, leading to a logical error in the first step of reasoning. Example is shown in Figure~\ref{fig: Error Sample for Logical Error}

\begin{figure*}[t]
\centering
\includegraphics[width=0.9\textwidth]{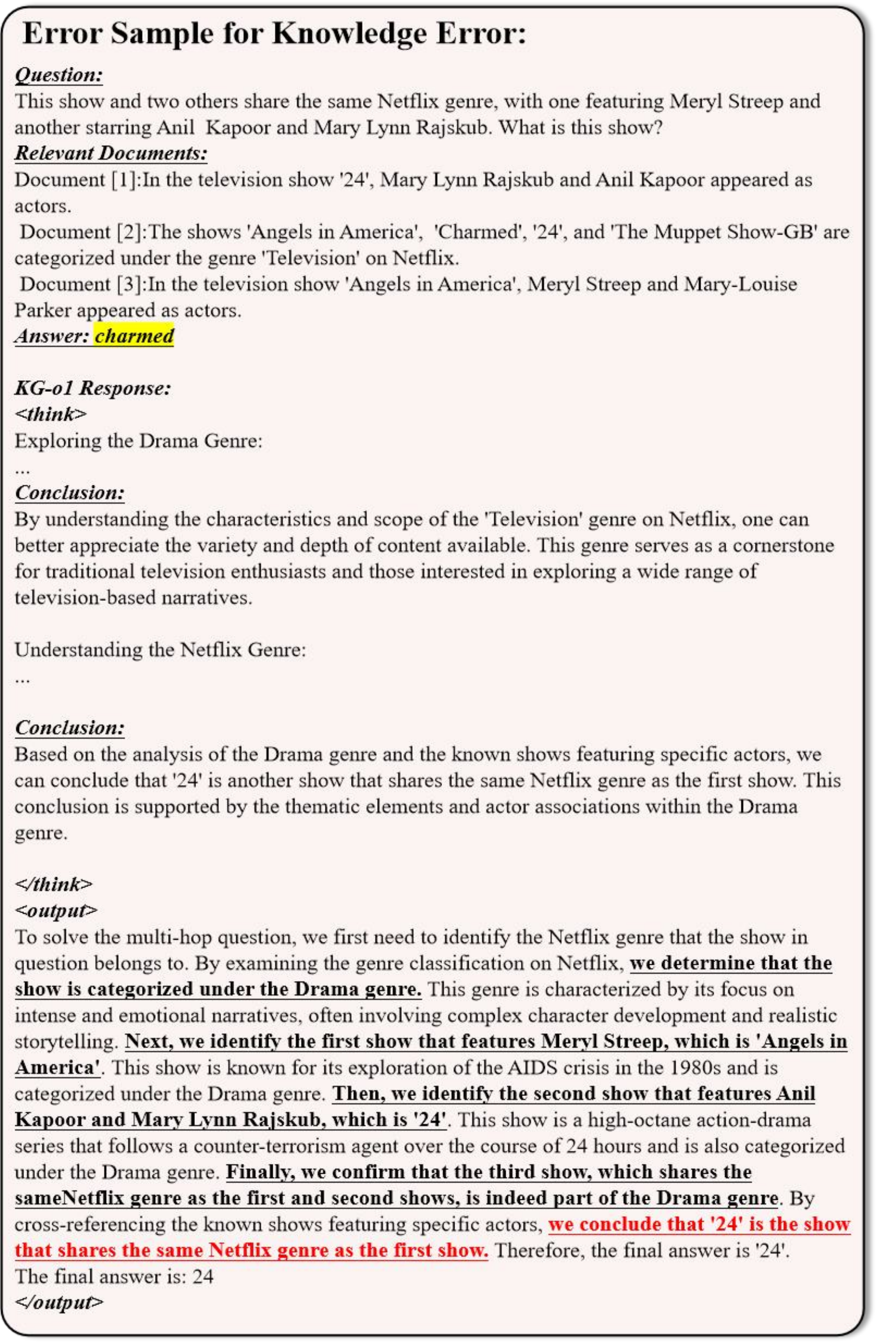} 
\caption{Sample for Knowledge Error}
\label{fig: Error Sample for Knowledge Error}
\end{figure*}

\textbf{Knowledge error}: This type of error arises when models incorrectly select irrelevant knowledge or incorrect relevant knowledge as the final answer, leading to reasoning errors. Such mistakes typically happen when the reasoning step involves a high density of knowledge.
For example, consider the question:\textit{This show and two others share the same Netflix genre, with one featuring Meryl Streep and another starring Anil Kapoor and Mary Lynn Rajskub. What is this show?}
The model correctly identifies the other two shows as Angels in America and 24, but in the final step, it selects incorrect knowledge, leading to an error in the reasoning process. Example is shown in Figure~\ref{fig: Error Sample for Knowledge Error}

\end{document}